\pdfoutput=1
\documentclass{article}
\PassOptionsToPackage{numbers,square,sort&compress}{natbib}
\usepackage[utf8]{inputenc}
\usepackage{authblk}
\usepackage{main}
\usepackage{microtype}
\usepackage{graphicx}
\usepackage{subfig}
\usepackage{times}
\usepackage{latexsym}
\usepackage{amsmath,amsfonts,amssymb,amsthm}
\usepackage{float}
\usepackage{placeins}
\usepackage{footnote}
\usepackage{enumitem}
\usepackage{bm}
\usepackage{arydshln}
\usepackage{booktabs}
\usepackage{multicol}
\usepackage{multirow}
\usepackage{color}
\usepackage{xcolor}     
\usepackage{colortbl}
\usepackage[table]{xcolor} %
\definecolor{royalblue}{RGB}{65,105,225}
\definecolor{purple}{RGB}{128,0,128}
\definecolor{selfevolagent_dark}{HTML}{37D2A6} 
\definecolor{selfevolagent_light}{HTML}{9BE9D3}
\definecolor{selfevolagent_lighter}{HTML}{CDF4E9}

\usepackage{bbding}
\usepackage{makecell}
\usepackage{mathtools}
\usepackage{imakeidx}
\usepackage{longtable}
\usepackage{wrapfig}
\makeindex
\usepackage{arydshln}
\usepackage{lipsum}
\usepackage{natbib}
\usepackage[edges]{forest}
\usepackage[normalem]{ulem}
\definecolor{mydarkblue}{rgb}{0,0.08,0.45}
\usepackage[colorlinks=true,linkcolor=mydarkblue,citecolor=mydarkblue,filecolor=mydarkblue,urlcolor=mydarkblue]{hyperref}
\usepackage{caption}
\usepackage{CJKutf8}
\usepackage{awesomebox} %
\usepackage{bbding}
\usepackage[most]{tcolorbox}
\usepackage{caption}

\usepackage[tikz]{bclogo}
\usepackage[framemethod=tikz]{mdframed}
\definecolor{bgblue}{RGB}{245,243,253}
\definecolor{ttblue}{RGB}{91,194,224}
\usetikzlibrary{positioning, shapes.geometric, shapes, arrows.meta}
\usepackage{tabularx}
\usepackage{geometry}
\usepackage{pifont}
\definecolor{myblue}{RGB}{0,80,160}
\definecolor{myyellow}{RGB}{210,170,0}
\definecolor{mygray}{RGB}{100,100,100}
\definecolor{mypurple}{RGB}{120,80,160}

\mdfdefinestyle{mystyle}{%
  rightline=true,
  innerleftmargin=10,
  innerrightmargin=10,
  outerlinewidth=3pt,
  topline=false,
  rightline=true,
  bottomline=false,
  skipabove=\topsep,
  skipbelow=\topsep
}

\newtcolorbox{myboxi}[1][]{
  breakable,
  title=#1,
  colback=red!5,
  colbacktitle=red!5,
  coltitle=black,
  fonttitle=\bfseries,
  bottomrule=0pt,
  toprule=0pt,
  leftrule=2pt,
  rightrule=2pt,
  titlerule=0pt,
  arc=0pt,
  outer arc=0pt,
  colframe=red,
}

\newtcolorbox{myboxnote}[1][]{
  breakable,
  title=#1,
  colback=orange!0,
  colbacktitle=orange!0,
  coltitle=black,
  fonttitle=\bfseries,
  bottomrule=0pt,
  toprule=0pt,
  leftrule=2pt,
  rightrule=2pt,
  titlerule=0pt,
  arc=0pt,
  outer arc=0pt,
  colframe=orange,
}

\newtcolorbox{myboxii}[1][]{
  breakable,
  freelance,
  title=#1,
  colback=white,
  colbacktitle=white,
  coltitle=black,
  fonttitle=\bfseries,
  bottomrule=0pt,
  boxrule=0pt,
  colframe=white,
  overlay unbroken and first={
  \draw[red!75!black,line width=3pt]
    ([xshift=5pt]frame.north west) -- 
    (frame.north west) -- 
    (frame.south west);
  \draw[red!75!black,line width=3pt]
    ([xshift=-5pt]frame.north east) -- 
    (frame.north east) -- 
    (frame.south east);
  },
  overlay unbroken app={
  \draw[red!75!black,line width=3pt,line cap=rect]
    (frame.south west) -- 
    ([xshift=5pt]frame.south west);
  \draw[red!75!black,line width=3pt,line cap=rect]
    (frame.south east) -- 
    ([xshift=-5pt]frame.south east);
  },
  overlay middle and last={
  \draw[red!75!black,line width=3pt]
    (frame.north west) -- 
    (frame.south west);
  \draw[red!75!black,line width=3pt]
    (frame.north east) -- 
    (frame.south east);
  },
  overlay last app={
  \draw[red!75!black,line width=3pt,line cap=rect]
    (frame.south west) --
    ([xshift=5pt]frame.south west);
  \draw[red!75!black,line width=3pt,line cap=rect]
    (frame.south east) --
    ([xshift=-5pt]frame.south east);
  },
}

\usepackage{fancyhdr} %
\usepackage{blindtext} %

\fancypagestyle{mwmcover}{%
  \fancyhf{}%
  \fancyhead[L]{%
    \includegraphics[height=0.42in]{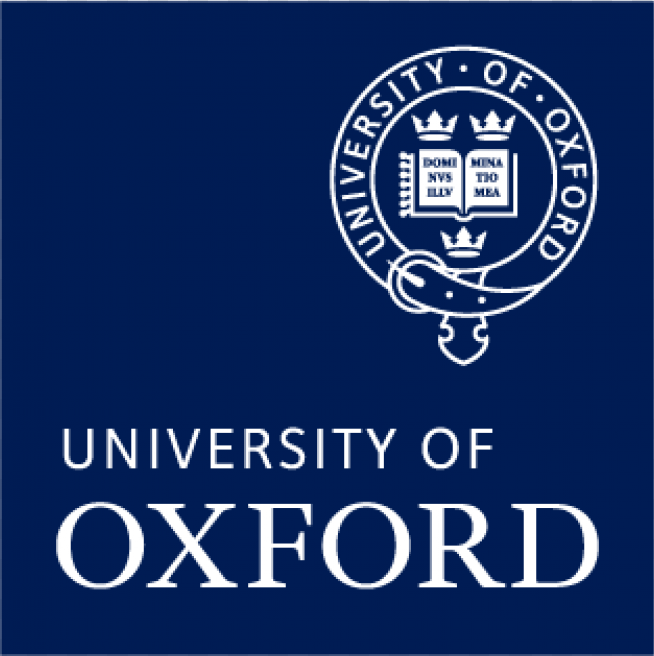}%
    \hspace{0.14in}%
    \includegraphics[height=0.42in]{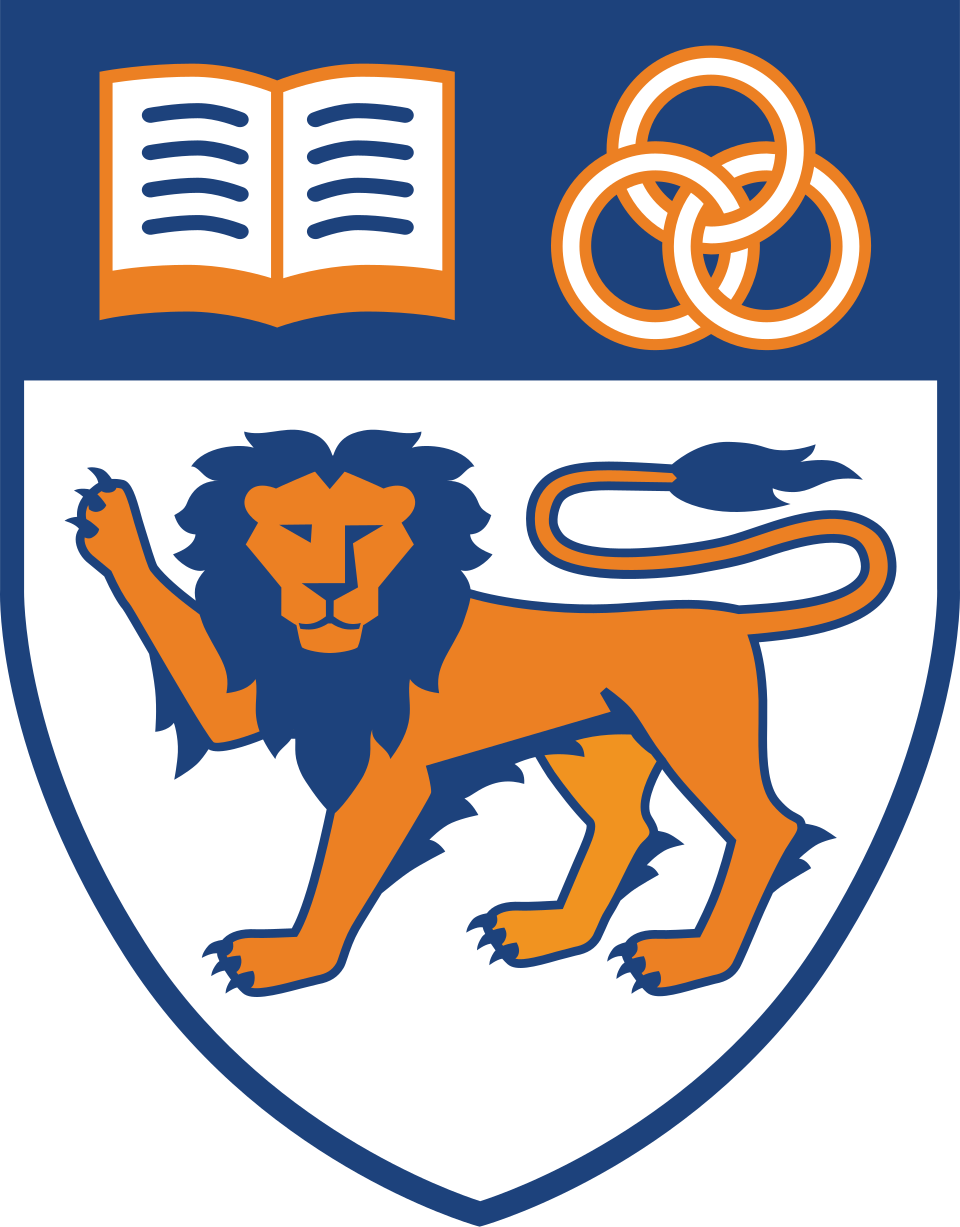}%
  }%
  \fancyhead[R]{\includegraphics[height=0.42in]{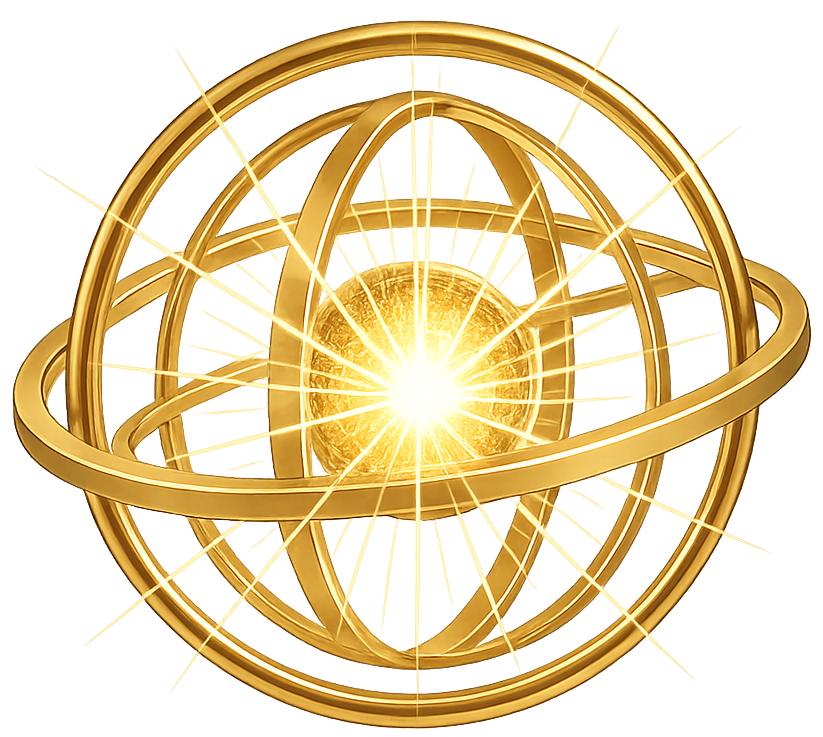}}%
  \renewcommand{\headrulewidth}{0pt}%
  \fancyfoot[C]{\thepage}%
}

\usepackage{forest}
\usepackage{xcolor}
\usepackage{tikz}
\usepackage{setspace}
\definecolor{lightblue}{RGB}{173,216,230}
\definecolor{lightgreen}{RGB}{144,238,144}
\definecolor{lightyellow}{RGB}{255,255,224}
\definecolor{lightpink}{RGB}{255,182,193}
\definecolor{hidden-draw}{RGB}{0,0,0}

\DeclareCaptionFont{black}{\color{black}}

\definecolor{myblue}{rgb}{0.9, 0.1, 0.94}
\definecolor{mygreen}{rgb}{0.64, 0.56, 0.88}
\definecolor{myyellow}{rgb}{0.68, 0.6, 0.1}
\definecolor{fancygreen}{rgb}{0.33, 0.68, 0.20}
\definecolor{salmon}{rgb}{0.94, 0.52, 0.49}
\definecolor{tablegreen}{rgb}{0.82, 0.94, 0.75}
\definecolor{tableblue}{rgb}{0.81, 0.90, 0.94}
\definecolor{tablered}{rgb}{0.97, 0.85, 0.85}
\definecolor{tableorange}{rgb}{0.96, 0.85, 0.81}

\newenvironment{itemize*}%
 {\leftmargini=10pt\begin{itemize}%
  \setlength{\itemsep}{0pt}%
  \setlength{\parskip}{0pt}%
  }%
 {\end{itemize}}
\newenvironment{enumerate*}%
 {\begin{enumerate}%
  \setlength{\itemsep}{0pt}%
  \setlength{\parskip}{0pt}}%
 {\end{enumerate}}

\usepackage{xcolor}
\usepackage{listings}

\newcommand\JSONnumbervaluestyle{\color{blue}}
\newcommand\JSONstringvaluestyle{\color{red}}

\newif\ifcolonfoundonthisline

\makeatletter

\lstdefinestyle{json}
{
  showstringspaces    = false,
  keywords            = {false,true},
  alsoletter          = 0123456789.,
  morestring          = [s]{"}{"},
  stringstyle         = \ifcolonfoundonthisline\JSONstringvaluestyle\fi,
  MoreSelectCharTable =%
    \lst@DefSaveDef{`:}\colon@json{\processColon@json},
  basicstyle          = \ttfamily,
  keywordstyle        = \ttfamily\bfseries,
}

\newcommand\processColon@json{%
  \colon@json%
  \ifnum\lst@mode=\lst@Pmode%
    \global\colonfoundonthislinetrue%
  \fi
}

\lst@AddToHook{Output}{%
  \ifcolonfoundonthisline%
    \ifnum\lst@mode=\lst@Pmode%
      \def\lst@thestyle{\JSONnumbervaluestyle}%
    \fi
  \fi
  \lsthk@DetectKeywords%
}

\lst@AddToHook{EOL}%
  {\global\colonfoundonthislinefalse}

\makeatother

\usepackage{etoolbox}
\usepackage{natbib}
\setcitestyle{numbers,square,comma,sort&compress}

\usepackage{url}

\newtheorem{property}{Property}[section]

\newtheorem*{definition*}{Definition}

\definecolor{mwmblue}{HTML}{E8F3FF}
\definecolor{mwmblueframe}{HTML}{2F6DAE}
\definecolor{mwmgreen}{HTML}{EAF8EF}
\definecolor{mwmgreenframe}{HTML}{2E8B57}
\definecolor{mwmamber}{HTML}{FFF4D6}
\definecolor{mwmamberframe}{HTML}{B88700}
\definecolor{mwmred}{HTML}{FFF0F0}
\definecolor{mwmredframe}{HTML}{B03030}
\definecolor{mwmviolet}{HTML}{F3EEFF}
\definecolor{mwmvioletframe}{HTML}{6F4BB7}

\newtcolorbox{mwmquotebox}[1][]{
  breakable,
  colback=mwmblue,
  colframe=mwmblueframe,
  boxrule=0.5pt,
  arc=2mm,
  left=6pt,
  right=6pt,
  top=5pt,
  bottom=5pt,
  #1
}

\newtcolorbox{mwmdefinitionbox}[1][]{
  breakable,
  colback=mwmgreen,
  colframe=mwmgreenframe,
  boxrule=0.7pt,
  arc=2mm,
  left=6pt,
  right=6pt,
  top=5pt,
  bottom=5pt,
  title=#1
}

\newtcolorbox{mwminsightbox}[1][]{
  breakable,
  colback=mwmamber,
  colframe=mwmamberframe,
  boxrule=0.7pt,
  arc=2mm,
  left=6pt,
  right=6pt,
  top=5pt,
  bottom=5pt,
  title=#1
}

\newtcolorbox{mwmplaceholderbox}[1][]{
  breakable,
  colback=mwmred,
  colframe=mwmredframe,
  boxrule=0.7pt,
  arc=2mm,
  left=6pt,
  right=6pt,
  top=5pt,
  bottom=5pt,
  title=#1
}

\definecolor{mwmfindnavy}{RGB}{46,71,135}
\definecolor{mwmfindback}{RGB}{243,245,250}
\definecolor{mwmclaim}{RGB}{168,24,24}
\newcommand{\coreclaim}[1]{\textbf{\textcolor{mwmclaim}{#1}}}
\newtcolorbox{mwmfindingbox}{
  colback=mwmfindback,
  colframe=mwmfindnavy,
  boxrule=0pt,
  leftrule=2.2pt,
  arc=1mm,
  left=7pt,
  right=7pt,
  top=4.5pt,
  bottom=4.5pt
}

\newcommand{\target}{\epsilon}
\newcommand{\Mentis}{\textsc{Mentis}}

\makeatletter
\newcommand{\appendixtocname}{Appendix Contents}
\newcommand{\appendixtoc}{%
  \begingroup
  \setcounter{tocdepth}{2}%
  \par\noindent{\bfseries\Large\appendixtocname}\par
  \vspace{0.6em}%
  \@starttoc{apc}%
  \endgroup
  \vspace{0.6em}%
}
\newcommand{\startappendixtoccapture}{%
  \let\StdSection\section
  \renewcommand\section[1]{%
    \StdSection{##1}%
    \addcontentsline{apc}{section}{\protect\numberline{\thesection}##1}%
  }%
  \let\StdSubsection\subsection
  \renewcommand\subsection[1]{%
    \StdSubsection{##1}%
    \addcontentsline{apc}{subsection}{\protect\numberline{\thesubsection}##1}%
  }%
}
\makeatother

\begin{document}

\title{Mental World Modeling}

\author{
    \centering
    Hao Fei\textsuperscript{1}
    \thanks{Corresponding author. Email: \texttt{haofei7419@gmail.com}.}
    \quad
    Yiran Zhao\textsuperscript{2}
    \thanks{Work done during internship at NUS.}
   \vspace{-0.2cm}
}
\vspace{-0.4cm}
\affil{\textsuperscript{1} University of Oxford \quad \textsuperscript{2}National University of Singapore }

\maketitle
\thispagestyle{mwmcover}
\setlength{\headsep}{0.3mm}
\vspace{-8mm}
\begin{center}
    \small\textbf{Homepage:}~\href{https://mental-world.github.io}{\texttt{https://mental-world.github.io}}
\end{center}
\vspace{1mm}

\begin{abstract}
World models enable a predictive substrate for planning and action, yet existing formulations merely answer a physical question: what/where it is, and how will it evolve. Human behavior, however, is driven by hidden mental state (what a person believes, wants, intends, feels, and considers socially permissible), so a model that tracks the physical scene but not \emph{what each agent knows and believes about it} predicts the wrong action for the right-looking scene. We formulate \textbf{Mental World Modeling} (MWM), a generic theoretical framework that makes mental variables core components of a world model rather than post-hoc rationales: MWM maintains a coupled physical-mental world state, renders a target-specific partial observation, and simulates how candidate actions jointly update both components. We instantiate the framework in \Mentis, a training-free and fully inspectable baseline that decomposes the process into state parsing, target-observation generation, action decomposition, coupled physical and mental transition, and branch-level value evaluation. On a manually constructed, quality-controlled dataset of situated decision scenarios spanning text, image, and sounding-video stories, experiments with 8 modern LLM-based world models demonstrate that explicitly modeling the mental state is essential for predicting human decisions. Deeper analyses further expose the bottlenecks of current mental world modeling. We expect MWM as a next stage of world modeling, from simulating physical scenes to simulating the minds that act in them.
\end{abstract}

\begin{figure}[h]
\vspace{-3mm}
    \centering
    \includegraphics[width=.98\linewidth]{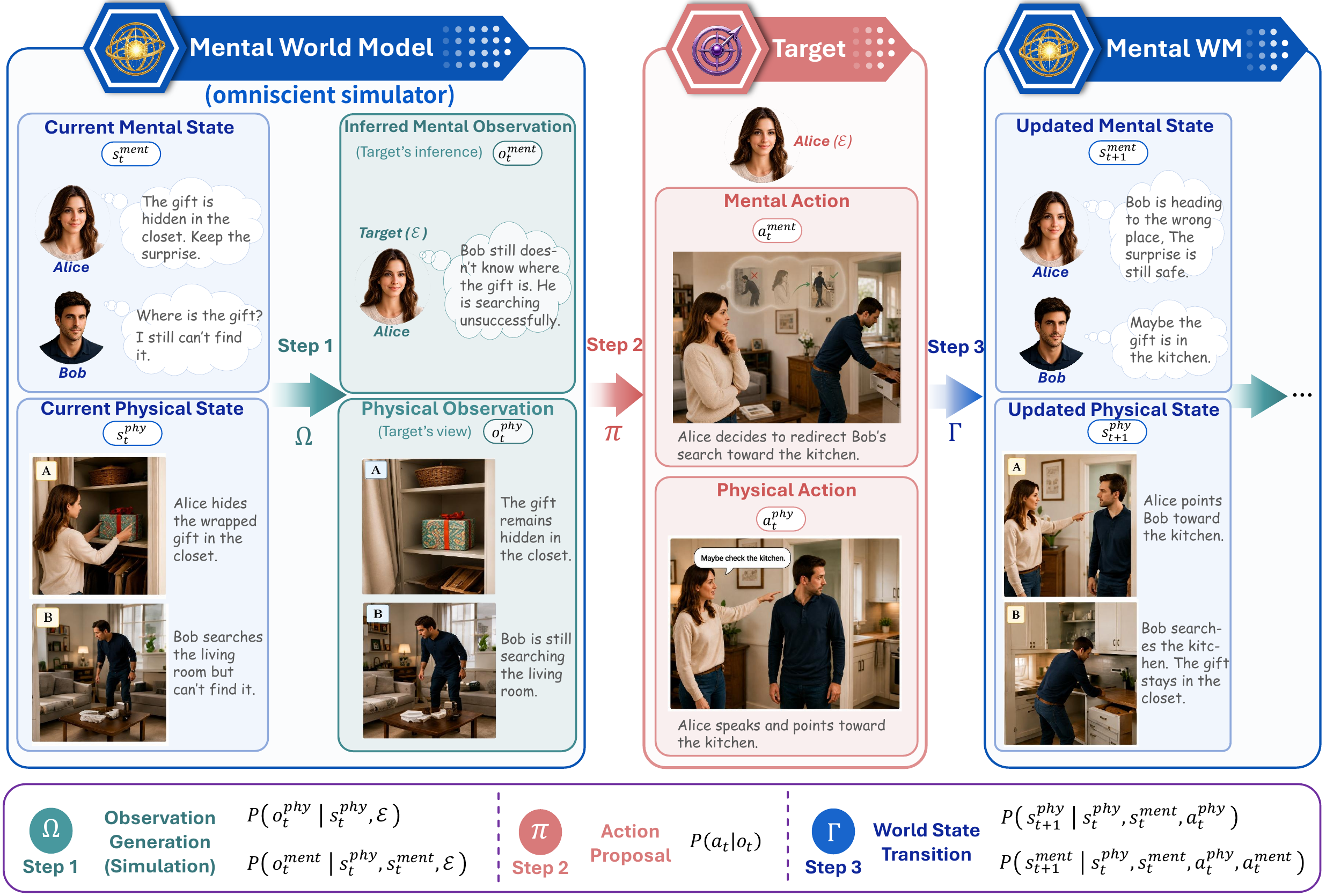}
\vspace{-2mm}
    \caption{\textbf{Mental World Modeling (MWM)} represents a scene as a coupled physical and mental world state, renders a target-specific observation, predicts candidate target actions, and simulates the next physical and mental states. Unlike physical world modeling alone, MWM explicitly tracks the unobserved beliefs, goals, intentions, emotions, relations, and norms that shape what the target agent will actually do.
    }
    \label{fig:teaser}
\end{figure}

\newpage
\pagestyle{fancy}
\lhead{\rightmark}
\renewcommand{\headrulewidth}{0.7pt}
\setlength{\headsep}{5mm}
\clearpage

\onecolumn
\begin{spacing}{1.1}
\tableofcontents
\end{spacing}
\setcounter{tocdepth}{3}%
\newpage

\section{Introduction}
\label{sec:intro}

\begin{mwmquotebox}
\emph{``If the organism carries a `small-scale model' of external reality and of its own possible actions within its head, it is able to try out various alternatives.''}
\par\smallskip\noindent\hfill\textit{Kenneth Craik, 1943~\citep{craik1943nature}}
\end{mwmquotebox}

Large language models (LLMs)~\citep{brown2020language,openai2023gpt4} now write programs, use tools, compose plans, and serve as the cognitive core of autonomous agents~\citep{yao2022react,shen2023hugginggpt,wang2023plan,wu2024next,wang2024voyager,park2023generative}, yet they do not by themselves constitute agents that live in, predict, and intervene on an evolving world. A capable agent must know what is true now, what would happen to itself if an action were taken, and how the world would change before the action is executed. This pressure has returned \textit{world models} to the center of AI research: a world model gives an agent a predictive substrate for counterfactual simulation, planning, and decision making~\citep{craik1943nature,ha2018world,lecun2022path}.

By now, the term \textit{world model} is still used in several ways, but three technological families are especially relevant here. \textit{Representation world models} learn compact latent states that support prediction and control, compressing observations into state abstractions and learning transition dynamics in representation space; this line runs from early recurrent world models through the Dreamer family to joint-embedding predictive architectures (JEPA)~\citep{ha2018world,hafner2023dreamerv3,lecun2022path}. \textit{Video-generative world models} synthesize temporally coherent future observations, with systems such as Sora, Genie, and Genie~3 as visible examples of learned world simulation~\citep{openai2024sora,bruce2024genie,deepmind2025genie3}. \textit{3D interactive world models} put spatial structure, navigability, editing, and persistent 3D environments at the center; Marble, for instance, is presented as a multimodal world model for generating, editing, expanding, and exporting 3D worlds~\citep{worldlabs2025marble}. These families differ in representation and interface, and each marks real progress toward spatially and physically grounded intelligence. But they share a default object of modeling: the physical substrate of the world, that is, objects, positions, geometry, motion, contact, occlusion, affordances, and perceptual continuity. They ask what the world is like and how the physical scene may evolve, and they leave the agents in that world as just another kind of moving object.

For human-centered intelligence, this is not enough. Human behavior is produced by the interaction between an external environment and an internal mental-social configuration. A service robot must infer whether a user is confused, impatient, or asking for help indirectly; a medical assistant must account for a patient's beliefs, risk perception, fear, and trust; a collaborative agent must recognize when a norm, role, or relationship makes a physically possible action socially inappropriate. These cases require tracking what agents know, believe, attend to, want, intend, feel, and regard as permissible. Cognitive science has studied these abilities through mental models, Theory of Mind (ToM), BDI agency, and embodied cognition~\citep{johnsonlaird1983mental,premack1978chimpanzee,bratman1987intention,varela1991embodied}, but most AI work either builds physical world models without mental state, or evaluates mental reasoning as isolated Theory-of-Mind question answering~\citep{sap2019socialiqa,chen2024tombench,ullman2023llm}. Neither view is sufficient for a world whose next state is jointly physical and mental.

\begin{figure}[!t]
    \centering
    \includegraphics[width=\linewidth]{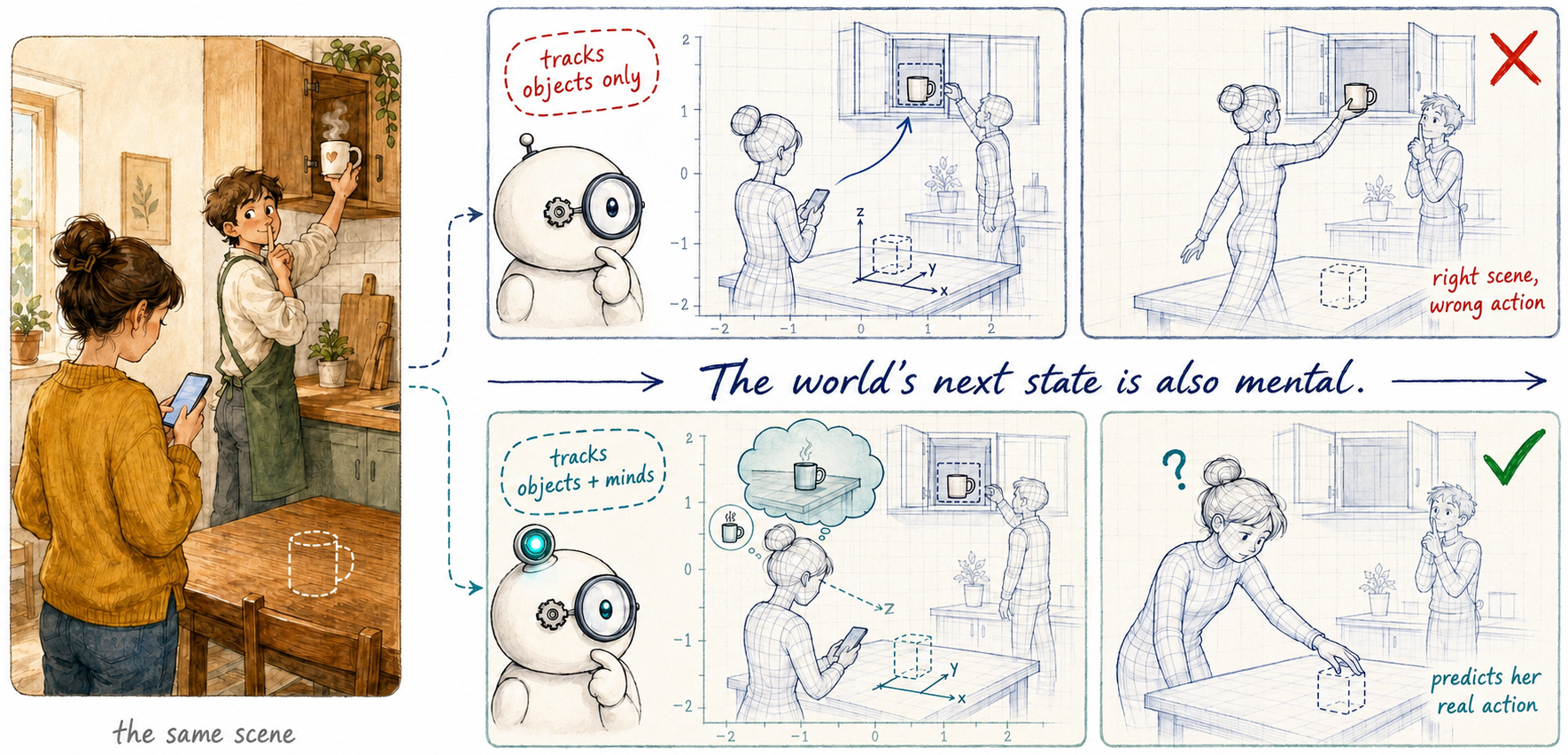}
    \caption{\textbf{The world's next state is also mental.} Her mug is moved
    to the cabinet while she looks away. A world model that tracks only objects
    predicts the wrong action for the right-looking scene (top); one that also
    tracks her belief predicts what she will actually do (bottom).}
    \label{fig:intro}
\end{figure}

This paper tries to formulate \textbf{Mental World Modeling} (MWM), a generic framework for building world models that jointly represent physical and mental dynamics. The goal is not to simulate a person's private conscious experience. Instead, MWM defines an external, approximate, task-relevant world simulator whose global state contains both physical variables and mental variables.
As illustrated in Figure \ref{fig:teaser}, a target agent observes only a partial, first-person rendering of this global state: what the target sees, hears, knows, believes, attends to, wants, feels, and considers socially permissible. The target agent then acts from this observation, while the world model predicts how the action changes both the material scene and the mental-social configuration. Formally, we build on the logic of partially observable Markov decision processes (POMDPs)~\citep{kaelbling1998planning}, where MWM maintains a joint state $s_t=(s_t^{\mathrm{phy}},s_t^{\mathrm{ment}})$, renders a target-specific observation $o_t^{\target}$, decomposes action into physical and mental aspects $a_t=(a_t^{\mathrm{phy}},a_t^{\mathrm{ment}})$, and defines an action-conditioned transition over the next joint state. This framing makes the central claim explicit: a world model for social intelligence must simulate not only \textit{what will happen}, but also \textit{what the target thinks is happening and why the target would act}.

To make the framework testable, we implement \Mentis, a modular reference implementation. Given a situated scene, a target agent, and a set of candidate actions, it parses the scene into the typed physical-mental state, renders the target's partial observation from that state, decomposes each candidate into its physical and mental aspects, simulates the coupled successor state of every branch, and scores each branch for physical plausibility, mental consistency, and social appropriateness before a deterministic rule selects the action. 
\Mentis{} is training-free, so its results reflect the modeling structure rather than fitted parameters. Every stage emits a machine-checkable artifact, so states, observations, and simulated futures can be logged, judged against annotations, or replaced by gold values. 

We evaluate on a manually constructed, quality-controlled data of situated decision records spanning text, image, and sounding-video realistic stories, each annotated with the full gold process, state and observation, and actions. 
A series of experiment groups test the framework where it could fail, and their results agree. The central claim is verified first, i.e., explicit mental world modeling is necessary for predicting human decisions. Full MWM is the best configuration for all 8 LLM-based world models we test, removing the mental channel degrades every one of them, and the gains are largest exactly on interpersonal scenes, where hidden mental variables carry the decision. 
Beyond necessity, oracle interventions turn the remaining human gap into a diagnosis, tracing most of it to transition simulation, the stage that predicts how the coupled world changes, and thereby setting the agenda for improving MWM systems.

To sum up, this paper contributes by:
\begin{itemize}[leftmargin=1.3em,itemsep=2pt,topsep=2pt,parsep=0pt]
    \item proposing \textbf{Mental World Modeling}, a formal mathematical framework that unifies physical and mental dynamics within a single world-modeling view, enabling AI systems to reason about not only how environments change, but also how agents perceive, interpret, and mentally respond to those changes.
    \item introducing \Mentis, a training-free benchmark baseline implementation that operationalizes MWM through state parsing, observation generation, action decomposition, coupled physical and mental transition, and branch-level value evaluation.
    \item conducting a systematic empirical study whose findings distill into two core claims: explicit mental world modeling is necessary for predicting human decisions, above all in social and interpersonal scenes; and the binding bottleneck of current MWM lies in transition simulation, which points the way for future improvement.
\end{itemize}

\section{Related Work and Preliminaries}
\label{sec:related}

\subsection{World Models}
\label{sec:related-world-models}

The term \textit{world model} commonly denotes a computational model that represents the current environment state and predicts how that state evolves. The concept has roots in cognitive accounts of internal simulation~\citep{craik1943nature}, but in contemporary AI it has become a practical architecture for prediction, planning, embodied control, and environment generation. The field is not unified around a single definition. 
To our knowledge, existing work can be separated into three main families according to the object being predicted; recent surveys provide more detail~\citep{zidan2026worldmodels}.

\paragraph{Representation world models.}
This line learns compact latent states that encode environment structure and dynamics. Early neural world models decompose perception, memory, and control so that an agent can learn behavior inside a predicted latent environment~\citep{ha2018world}. Dreamer-style agents learn latent dynamics models for planning and policy learning across diverse domains~\citep{hafner2023dreamerv3}. JEPA-like predictive architectures move further away from pixel reconstruction: they learn useful abstractions by predicting missing or future information in representation space rather than generating every perceptual detail~\citep{lecun2022path}. Recent systems scale this recipe further: V-JEPA 2 pretrains action-conditionable latent predictors on internet-scale video and transfers them to zero-shot robot planning~\citep{meta2025vjepa2}, and Dreamer 4 trains agents entirely inside its learned world model~\citep{hafner2025dreamer4}. The strength of this family is abstraction. Its limitation, from the viewpoint of MWM, is that the represented state is still typically a physical or perceptual state: objects, spatial relations, motion, affordances, and action consequences.

\paragraph{Video-generative world models.}
A second family treats world modeling as explicit synthesis of future observations. Video models are attractive because video naturally encodes temporal change, object persistence, motion, and approximate physical regularities. Sora frames video generation as a route toward general-purpose world simulation~\citep{openai2024sora}; Genie learns action-controllable interactive environments from unlabeled videos~\citep{bruce2024genie}; Genie 3 pushes this direction toward real-time navigable generated worlds~\citep{deepmind2025genie3}. These models make the physical and visual sense of ``world'' concrete: the model can generate what a future scene may look like, sometimes conditioned on user or agent actions. The family is advancing quickly toward physical fidelity and interactivity: Sora 2 adds synchronized audio and more accurate physics~\citep{openai2025sora2}; Cosmos, GAIA-2, and OmniDreams build world simulators for physical AI and closed-loop driving~\citep{nvidia2025cosmos,wayve2025gaia2,nvidia2026omnidreams}; and Matrix-Game 3.0 streams real-time interactive worlds with long-horizon memory~\citep{skywork2026matrixgame}. However, video generation mainly renders visible or spatially implied consequences. It does not, by default, maintain explicit variables for what an agent falsely believes, privately intends, emotionally feels, or socially ought to do.

\paragraph{3D interactive and spatial world models.}
A third family should be distinguished from video-centric generation. These systems model worlds as spatial structures that can be navigated, edited, expanded, exported, or used in downstream simulators. The core object is not merely a future frame sequence, but a persistent 3D or 4D environment with geometry, layout, camera control, and interaction affordances. Marble from World Labs is a representative recent example: it is described as a multimodal world model that creates 3D worlds from text, images, video, or coarse 3D layouts, and supports editing, expansion, composition, and export to 3D formats~\citep{worldlabs2025marble}. HunyuanWorld 1.0 and HY-World 2.0 likewise generate semantically layered, navigable 3D scenes from text, images, or video~\citep{tencent2025hunyuanworld,tencent2026hyworld2}, a line surveyed in~\citep{kong2025survey3d4d}. This family is especially relevant for embodied AI because it brings world models closer to explorable spatial environments. Still, its central variables remain spatial and physical. A navigable 3D apartment can encode where the kitchen is and how a user moves through it; it does not necessarily encode who believes the medicine is in the kitchen, who is anxious about taking it, or which social norm makes one action inappropriate.

While these families are complementary and important, where they give AI increasingly strong models of physical, visual, and spatial structure, the gap addressed by this paper is dinstinct: a real-world world model must also represent the latent mental and normative states that make human behavior intelligible. Physical dynamics can explain how a door opens; they do not explain why someone refuses to open it, whether they know what is behind it, or how opening it changes trust, embarrassment, obligation, or conflict.

\subsection{Cognitive, Mental, and Theory-of-Mind Modeling}
\label{sec:related-mental}

The closest intellectual foundations for MWM come from cognitive science, philosophy of action, developmental psychology, and multi-agent AI. Mental-model theory argues that people reason by constructing internal representations of situations and possible events, rather than manipulating propositions alone~\citep{craik1943nature,johnsonlaird1983mental}. Theory of Mind studies the capacity to attribute beliefs, desires, intentions, knowledge, and other mental states to oneself and others~\citep{premack1978chimpanzee,wimmer1983beliefs,baroncohen1985autistic}. Affective computing treats emotions as variables a machine can represent, recognize, and respond to~\citep{picard1997affective}. These traditions are central to our framing because they identify the missing variables in physical world modeling: the world relevant to action is partly the world as an agent understands it.

In AI, one influential formal tradition is the Belief-Desire-Intention model of agency, which represents practical reasoning through what an agent believes, wants, and commits to doing~\citep{bratman1987intention,rao1995bdi}. Neural and multi-agent work on Machine Theory of Mind similarly trains models to infer other agents' hidden states from behavior~\citep{rabinowitz2018machine}. More recently, social reasoning and ToM benchmarks have tested whether language models can answer questions about motivation, belief, false belief, social commonsense, or interpersonal situations, including ToMi~\citep{le2019revisiting}, SocialIQA~\citep{sap2019socialiqa}, FANToM~\citep{kim2023fantom}, and ToMBench~\citep{chen2024tombench}, with SOTOPIA extending the evaluation to interactive social scenarios~\citep{zhou2024sotopia}. Other studies show that language-model ToM behavior can be brittle under small false-belief or task-format changes~\citep{kosinski2024evaluating,ullman2023llm}.

These lines are highly relevant, but their modeling unit is usually different from ours. Mental-model and ToM research often focuses on inferring or evaluating a mental state; benchmark work often reduces the problem to a static answer; BDI systems usually formalize an agent's own practical reasoning. MWM instead asks for a world model that contains both physical and mental state, renders a target agent's partial observation, accepts candidate actions, and simulates the next joint state. This shift matters because mental variables are not only labels to infer; they are state variables that evolve as physical events, speech acts, observations, and social consequences unfold.

\subsection{POMDP-based World Modeling}
\label{sec:related-pomdp-social}

POMDPs provide the classical formalism for decision making when the underlying state is only partially observed~\citep{kaelbling1998planning}. In a standard POMDP, an agent maintains a belief over hidden environment states, chooses actions from observation history, and receives observations generated from the latent state. This logic has directly shaped learned world models: PlaNet formulates image-based control as a POMDP, learns recurrent latent dynamics from pixels, infers hidden state from action-observation history, and plans online in latent space~\citep{hafner2019planet}; Dreamer-style agents follow the same broad state-space modeling pattern for policy learning and planning~\citep{hafner2023dreamerv3}. Multi-agent variants, such as Interactive POMDPs, further incorporate models of other agents into the state space~\citep{gmytrasiewicz2005framework}. More recently, language models themselves have been repurposed as world models: RAP treats the LLM as both reasoning agent and world model, and plans over its predicted states with Monte Carlo Tree Search~\citep{hao2023reasoning}. These methods are important precursors for MWM, as they show how latent state, partial observation, action, and transition can be unified; Section~\ref{sec:math-formalization} adopts exactly this scaffold. Their latent states, however, are usually optimized for physical control, reward prediction, visual dynamics, or nested decision-theoretic planning, rather than explicit physical-mental world simulation.

Recent relevant work on \textit{Social World Models} (SWM) similarly adopts a POMDP/Dec-POMDP-inspired view of social interaction~\citep{zhou2025socialworldmodels}. Its S3AP representation parses narratives into structured fields such as state, observations, actions, and mental states, improving LLM social reasoning and agent decision making. This is close to our motivation, but not identical in scope. SWM mainly proposes a structured social representation and LLM-powered reasoning framework for social narratives and interactions; MWM is a broader physical-mental world-modeling framework that treats target observation as a first-person rendering of a third-person joint state, and explicitly simulates branch-level physical and mental transitions in \Mentis.

\section{Theoretical Framework}
\label{sec:theory}

\subsection{Theoretical Grounding}
\label{sec:theory-grounding}

MWM is not a literal theory of consciousness. It is a modeling proposal, i.e., an external simulator should maintain the variables needed to predict social action and its consequences. The theoretical burden of this section is therefore to justify four commitments: (i) the world state should include both physical and mental variables; (ii) a target agent should receive a first-person partial observation rather than the full state; (iii) an action should have both a physical carrier and mental-social meaning; and (iv) the transition to the next state should be jointly physical and mental. These commitments follow from several well-established theories; Appendix~\ref{app:theory-grounding} summarizes the correspondence, pairing each theoretical source with the commitment it supports.

\vspace{2mm}
\begin{mwmquotebox}[title={Theory-of-Mind premise}, colback=mwmviolet, colframe=mwmvioletframe]
\emph{``An individual has a theory of mind if he imputes mental states to himself and others.''}
\hfill\textit{Premack and Woodruff~\citep{premack1978chimpanzee}}
\end{mwmquotebox}

\noindent\textbf{Mental state as a causal state variable.}
Theory of Mind and BDI agency justify treating beliefs, desires, intentions, knowledge, and affect as variables that help cause behavior, rather than as explanations added after behavior is observed~\citep{premack1978chimpanzee,wimmer1983beliefs,baroncohen1985autistic,bratman1987intention,rao1995bdi}. In the same physical room, two agents with different beliefs may take different actions; after the same utterance, two agents with different goals may update their mental states differently. Lake et al.~\citep{lake2017building}'s account of human-like learning makes the same point from the AI side: human-like systems need causal, compositional structure grounded in both intuitive physics and intuitive psychology. This motivates that the world state must carry a physical component (entities, relations, environment) and a mental component (beliefs, attention, goals, intentions, affect, norms, social relations), because both components change the distribution over future actions and future states. Section~\ref{sec:math-formalization} formalizes this factorization as $s_t=(s_t^{\mathrm{phy}},s_t^{\mathrm{ment}})$.

\vspace{2mm}
\begin{mwmquotebox}[title={Embodied cognition premise}, colback=mwmgreen, colframe=mwmgreenframe]
\emph{``Cognition depends upon the kinds of experience that come from having a body with various sensorimotor capacities.''}
\hfill\textit{Varela, Thompson, and Rosch~\citep{varela1991embodied}}
\end{mwmquotebox}

\noindent\textbf{Mental state is situated, not free-floating.}
Embodied and enactive cognition reject a view in which mental content can be interpreted independently of the agent's situated interaction with the world~\citep{varela1991embodied}. In MWM terms, a mental state is not a detached text label: its meaning depends on the physical scene. Disappointment depends on what was expected and what occurred; politeness depends on role, place, and norm; suspicion depends on what was visible, said, or hidden. Active inference offers a compatible formal picture: internal states and external states are conditionally separated by a Markov blanket of sensory and active states, and remain coupled only through perception and action~\citep{friston2010free}. For MWM, this is the conceptual reason to model observation and action as the two interfaces between physical and mental dynamics. A target's mental state is updated through sensory access to the world; the target's action changes the physical scene and, through others' perception of it, alters their beliefs, emotions, and obligations.

\vspace{2mm}
\begin{mwmquotebox}[title={Affordance premise}, colback=mwmamber, colframe=mwmamberframe]
\emph{``An affordance points both ways, to the environment and to the observer.''}
\hfill\textit{Gibson~\citep{gibson1979ecological}}
\end{mwmquotebox}

\noindent\textbf{Observation is already interpreted.}
Ecological psychology clarifies why MWM should not identify observation with raw perception~\citep{gibson1979ecological}. A target observes not only geometry, but also action possibilities and social meanings. A chair can be \textit{sit-able}, \textit{offer-able}, \textit{blocked}, or \textit{socially reserved}; these are not properties of pixels alone, but joint functions of the physical scene, the target's own mental state, its social relations, and the norms currently in force. This is why the same physical scene can render different observations for different agents. In MWM, the target observation $o_t^{\target}$ is therefore a first-person rendering of the global state, not a copy of it.

\vspace{2mm}
\noindent\textbf{Partial observability gives the formal scaffold.}
POMDPs formalize the gap between a latent world state and an agent's local observation~\citep{kaelbling1998planning}. MWM adopts this scaffold but changes what the latent state contains. The hidden state is not only an occluded object or unobserved physical variable; it may be a false belief, private intention, emotion, role relation, or social norm. The target acts from $o_t^{\target}$, while the world model maintains the richer state $s_t=(s_t^{\mathrm{phy}},s_t^{\mathrm{ment}})$ for simulation.

\begin{property}[Why a joint state is necessary]
Let $p^\star(a_{t}^{\target}\mid s_t)$ be the true target-action distribution over the joint state $s_t=(s_t^{\mathrm{phy}},s_t^{\mathrm{ment}})$. A physical-only representation, which retains only $s_t^{\mathrm{phy}}$, is insufficient whenever there exist mental components $s_t^{\mathrm{ment}}\neq\tilde{s}_t^{\mathrm{ment}}$ such that
\begin{equation}
p^\star\big(a_t^{\target}\mid s_t^{\mathrm{phy}},s_t^{\mathrm{ment}}\big) \neq p^\star\big(a_t^{\target}\mid s_t^{\mathrm{phy}},\tilde{s}_t^{\mathrm{ment}}\big).
\label{eq:physical-insufficiency}
\end{equation}
A mental-only representation, which retains only $s_t^{\mathrm{ment}}$, is insufficient whenever there exist physical components $s_t^{\mathrm{phy}}\neq\tilde{s}_t^{\mathrm{phy}}$ such that
\begin{equation}
p^\star\big(a_t^{\target}\mid s_t^{\mathrm{phy}},s_t^{\mathrm{ment}}\big) \neq p^\star\big(a_t^{\target}\mid \tilde{s}_t^{\mathrm{phy}},s_t^{\mathrm{ment}}\big).
\label{eq:mental-insufficiency}
\end{equation}
\end{property}

\begin{proof}[Proof sketch]
A model that observes only $s_t^{\mathrm{phy}}$ cannot distinguish the two states $(s_t^{\mathrm{phy}},s_t^{\mathrm{ment}})$ and $(s_t^{\mathrm{phy}},\tilde{s}_t^{\mathrm{ment}})$, so it must assign the same action distribution to both; Eq.~\eqref{eq:physical-insufficiency} states that the true distributions differ, so no physical-only predictor can be sufficient in such cases. The mental-only case follows symmetrically from Eq.~\eqref{eq:mental-insufficiency}. Social scenarios routinely instantiate both conditions: false-belief tasks share the same physical layout but differ in belief, while visibility and affordance tasks share similar mental goals but differ in physical access.
\end{proof}

\subsection{Core Definition}
\label{sec:core-definition}

\begin{table}[!t]
\caption{Notation used in the MWM formulation. 
}
\label{tab:mwm-notation}
\centering
\scriptsize
\renewcommand{\arraystretch}{1.08}
\rowcolors{2}{gray!7}{white}
\begin{tabularx}{\linewidth}{p{0.18\linewidth}p{0.28\linewidth}X}
\toprule
\rowcolor{white}
\textbf{Group} & \textbf{Notation} & \textbf{Meaning} \\
\midrule
Indices and agents & $t$; $i,j$; $N$; $\mathcal{I}$; $\target$ & Time step; agent indices; number of agents; set of agents; the current target agent. \\
Groups & $G$; $\mathcal{G}$ & A group-level mental entity and the set of such groups, when a scene contains collective actors. \\
State spaces & $\mathcal{S}$; $\mathcal{S}^{\mathrm{phy}}$; $\mathcal{S}^{\mathrm{ment}}$ & Joint world-state space; physical state space; mental and social state space. \\
World state & $s_t=(s_t^{\mathrm{phy}},s_t^{\mathrm{ment}})$; $s_{t+1}$ & Global third-person state maintained by the world model and its successor. \\
Physical state & $s_t^{\mathrm{phy}}=(O_t,C_t,R_t^{\mathrm{phy}},E_t)$ & Physical entities and scene conditions: objects, characters, physical relations, and environment. \\
Mental state & $s_t^{\mathrm{ment}}=(\{m_t^i\},\{m_t^G\},R_t^{\mathrm{ment}},\alpha_t)$ & Mental and social state: individual mental states, group mental states, mental relations, and atmosphere. \\
Individual mental fields & $\mathrm{id}_t^i,b_t^i,q_t^i,g_t^i,\iota_t^i,e_t^i,d_t^i,n_t^i,c_t^i$ & Identity, beliefs, attention, goals, intentions, emotions, dispositions, norms, and behavioral constraints of agent $i$. \\
Observation spaces & $\mathcal{O}^{\mathrm{phy}}$; $\mathcal{O}^{\mathrm{ment}}$ & Spaces of physical observations and mental observations. \\
Target observation & $o_t^{\target}=(o_t^{\target,\mathrm{phy}},o_t^{\target,\mathrm{ment}})$ & First-person partial observation rendered for the target agent. \\
Mental observation & $\omega_{\mathrm{self}}(\cdot)$; $\widehat{m}_{t,\target}^{j,(\ell)}$; $\ell$; $L$ & Self-observation operator; the target's $\ell$-order inference about agent $j$; ToM inference order; maximum ToM depth. \\
Action spaces & $\mathcal{A}^{\mathrm{phy}}$; $\mathcal{A}^{\mathrm{ment}}$ & Physical-action carrier space and mental or semantic action-content space. \\
Target action & $a_t^{\target}=(a_t^{\target,\mathrm{phy}},a_t^{\target,\mathrm{ment}})$ & Target action decomposed into physical carrier and mental or semantic content. \\
Observation functions & $\Omega_{\target}$; $\Omega_{\target}^{\mathrm{phy}}$; $\Omega_{\target}^{\mathrm{ment}}$ & Full target observation generator and its physical and mental components. \\
Perspective variables & $\kappa_t^{\target}$; $\rho_t^{\target}$ & Perceptual access variables and social-cognitive perspective variables used by observation generation. \\
ToM inference & $Q_{\target}^{(\ell)}$ & Target-specific distribution for $\ell$-order Theory-of-Mind inference. \\
Action proposal & $\Pi_{\target}$; $\Pi_{\target}^{\mathrm{phy}},\Pi_{\target}^{\mathrm{ment}}$; $\Pi_{\target}^{K}$; $K$ & Target policy over actions; its physical and mental factors; candidate-action generator; number of candidates. \\
Candidate actions & $\mathcal{A}_t^{\target}$; $a_{t,k}^{\target}$ & Candidate-action set proposed at time $t$ and its $k$-th member. \\
Transition model & $T_{\theta}$; $T_{\theta}^{\mathrm{phy}}$; $T_{\theta}^{\mathrm{ment}}$ & Joint transition kernel with parameters $\theta$, and its physical and mental factors. \\
Other agents & $a_t^{-\target}$ & Actions of non-target agents, marginalized in the target-centric formulation. \\
Branch evaluation & $\hat{s}_{t+1}^{k}$; $r_k$; $\hat{k}$ & Simulated successor state for candidate $k$; its branch score from the value evaluator; selected branch (instantiated by \Mentis{}, Section~\ref{sec:mentis}). \\
\bottomrule
\end{tabularx}
\rowcolors{2}{}{}
\renewcommand{\arraystretch}{1.0}
\end{table}

\begin{mwmdefinitionbox}[Definition (Mental World Modeling)]
\itshape
\textbf{Mental World Modeling} is a target-centric, action-conditioned world-modeling framework in which an external simulator (i) maintains a \emph{joint physical-mental world state}, (ii) renders a \emph{first-person partial observation} for a specified target agent, and (iii) simulates the \emph{next joint state} conditioned on the target's action. The state space factors as $\mathcal{S}=\mathcal{S}^{\mathrm{phy}}\times\mathcal{S}^{\mathrm{ment}}$, where $\mathcal{S}^{\mathrm{phy}}$ stores entities, relations, and environmental conditions, and $\mathcal{S}^{\mathrm{ment}}$ stores latent mental-social variables such as beliefs, attention, goals, intentions, emotions, preferences, norms, role relations, and atmosphere.
\end{mwmdefinitionbox}

Three clarifications delimit this definition.

\vspace{1mm}
\noindent\textbf{MWM is an objective simulator, not simulated consciousness.}
MWM does not claim to reproduce a person's private conscious experience. It is a modeling abstraction: the simulator maintains whatever variables are needed to predict what a target can observe, how the target may act, and how the scene evolves after that action. A physical world model asks what the environment is like and how it will change; an MWM additionally asks what the target \emph{thinks} the world is like, what the target wants, how target will interpret the others' actions and how others will interpret the target's action as well.

\vspace{1mm}
\noindent\textbf{The world model is not the agent.}
In many physical-control settings the learned world model is embedded inside the acting agent, so it is tempting to say that the model itself ``chooses'' the action. MWM separates these roles: the \emph{target agent} acts from an egocentric observation, while the \emph{world model} simulates consequences from a richer third-person state (Figure~\ref{fig:teaser}). The target may be wrong, uninformed, biased, or socially constrained; the world model still represents the hidden variables that explain why the target's observation differs from the global state. A cup handed across a table is one physical event, yet it may be an apology, a deception, an act of compliance, or a gesture of care, and only the world model holds the mental-social state that disambiguates it.

\vspace{1mm}
\noindent\textbf{One process, variable scope.}
The definition covers single-agent and multi-agent scenes with the same process. In a single-agent navigation problem, $\mathcal{S}^{\mathrm{ment}}$ may collapse to the target's goal and belief about the map; in a multi-agent social scene, it expands to beliefs, intentions, norms, interpersonal attitudes, group roles, and shared atmosphere. What changes is the dimensionality of the state, not the modeling law: a target-specific observation is rendered from a global state, the target acts from that observation, and the world model predicts a physical-mental successor state.

\begin{mwminsightbox}[Core modeling claim]
\itshape
The minimal object of simulation for social decision-making is not a physical trajectory alone and not a theory-of-mind answer alone. It is an action-conditioned transition over a coupled state: physical variables constrain what can happen, while mental variables determine what the same happening means to the agents involved.
\end{mwminsightbox}

\subsection{Formulation of Mental World Modeling}
\label{sec:math-formalization}

We formulate MWM as an augmented partially observable dynamic process. The augmentation is not cosmetic. A standard POMDP typically hides parts of the physical state. MWM treats mental and social variables as part of the latent world state itself, and therefore makes target-specific observation, action semantics, and mental-state transition first-class objects. 
Table~\ref{tab:mwm-notation} collects the symbols used throughout this section.

Figure~\ref{fig:teaser} illustrates the whole loop of the MWM process. The target agent receives a first-person observation by the MWM, produces an action, and the MWM then simulates the next joint state and the next observation.
Table~\ref{tab:mwm-basic-elements-overview} summarizes the high-level components of MWM.
$s_t$ is maintained by the world model; $o_t^{\target}$ is rendered from $s_t$ for a target agent; $a_t^{\target}$ is produced from that observation; the world model then simulates $s_{t+1}$ and the next observation. The remaining formulation makes each element precise.

\begin{table}[!t]
\caption{Basic elements of the MWM framework. Two computational roles interact through three coupled variables: world state, target observation, and target action.}
\label{tab:mwm-basic-elements-overview}
\centering
\small
\renewcommand{\arraystretch}{1.08}
\rowcolors{2}{gray!7}{white}
\begin{tabularx}{\linewidth}{p{0.16\linewidth}p{0.20\linewidth}p{0.24\linewidth}X}
\toprule
\rowcolor{white}
\textbf{Type} & \textbf{Element} & \textbf{Formal object} & \textbf{Role in MWM} \\
\midrule
& \textsc{TargetAgent}$_{\target}$
& $o_t^{\target}\mapsto a_t^{\target}$
& Situated actor; acts from a first-person partial observation. \\
\multirow{-2}{*}[-0.9em]{\makecell[l]{\textbf{Computational}\\\textbf{role}}}
& \textsc{WorldModel}
& $(s_t,a_t^{\target})\mapsto(s_{t+1},o_{t+1}^{\target})$
& External simulator; maintains state, applies action, predicts the successor state and observation. \\
\specialrule{0.85pt}{1.2mm}{1.2mm}
& World State
& $s_t=(s_t^{\mathrm{phy}},s_t^{\mathrm{ment}})$
& Third-person latent state: physical scene plus mental-social variables. \\
& Observation
& $o_t^{\target}=(o_t^{\target,\mathrm{phy}},o_t^{\target,\mathrm{ment}})$
& First-person rendering for the target; partial and perspective-dependent. \\
\multirow{-3}{*}{\makecell[l]{\textbf{Interaction}\\\textbf{variable}}}
& Action
& $a_t^{\target}=(a_t^{\target,\mathrm{phy}},a_t^{\target,\mathrm{ment}})$
& Coupled action: physical carrier plus mental or semantic content. \\
\bottomrule
\end{tabularx}
\rowcolors{2}{}{}
\renewcommand{\arraystretch}{1.0}
\end{table}

\subsubsection{Basic Elements}
\label{sec:mwm-basic-elements}

\paragraph{Roles.}
Let $\mathcal{I}=\{1,\ldots,N\}$ denote the set of agents or socially meaningful entities in a scene, and let $\target\in\mathcal{I}$ be the target whose next action is under consideration. MWM distinguishes two computational roles:
\begin{align}
\textsc{TargetAgent}_{\target}:&\quad
o_t^{\target} \mapsto a_t^{\target}, \label{eq:role-agent}\\
\textsc{WorldModel}:&\quad
\big(s_t,a_t^{\target}\big)
\mapsto
\big(s_{t+1},o_{t+1}^{\target}\big).
\label{eq:role-world-model}
\end{align}
The target agent is first-person and partial: it can only condition on its rendered observation $o_t^{\target}$. The world model is third-person and generative: it maintains a joint state $s_t$, receives the target action $a_t^{\target}$, and simulates the next state. Other agents need not be ignored. Their beliefs, goals, intentions, relations, and likely reactions are encoded in $s_t^{\mathrm{ment}}$ and can affect the transition even when the formulation is target-centric.

\providecommand{\taxleaf}[2]{\parbox{20.5em}{\raggedright\textbf{#1}\\ \emph{#2}}}

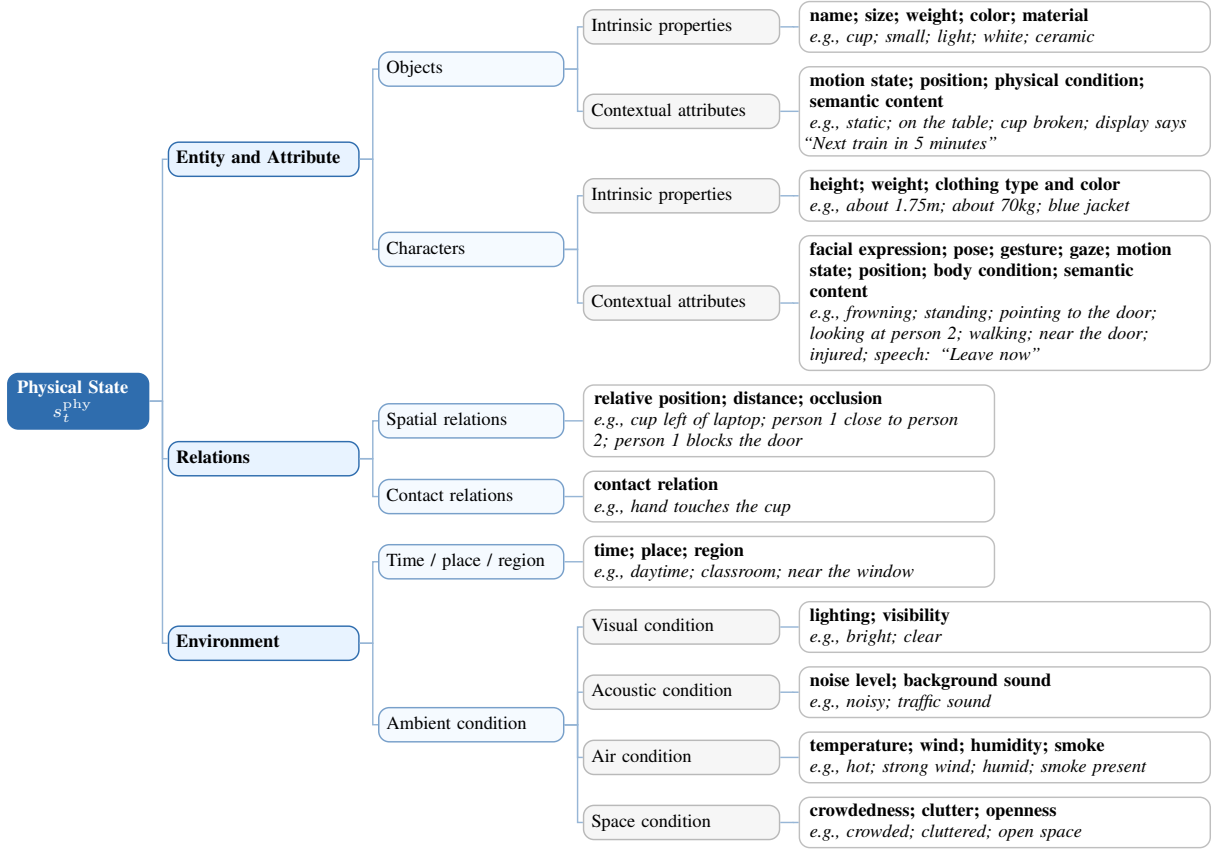
\begin{figure}[!t]
\centering
\scriptsize
\resizebox{\linewidth}{!}{%
\begin{forest}
forked edges,
for tree={
    grow=east,
    reversed=true,
    anchor=base west,
    parent anchor=east,
    child anchor=west,
    base=center,
    font=\scriptsize,
    rectangle,
    rounded corners,
    align=left,
    edge+={mwmblueframe!50, line width=0.55pt},
    inner xsep=3pt,
    inner ysep=2.5pt,
    s sep=5pt,
    l sep=7pt,
    line width=0.55pt,
},
where level=0{fill=mwmblueframe, draw=mwmblueframe, text=white, font=\bfseries\scriptsize, text width=6.8em}{},
where level=1{fill=mwmblue, draw=mwmblueframe, font=\bfseries\scriptsize, text width=9.5em}{},
where level=2{fill=mwmblue!45, draw=mwmblueframe!65, text width=9.2em}{},
where level=3{fill=gray!7, draw=gray!55, text width=9.8em}{},
where n children=0{fill=white, draw=gray!50, inner xsep=4pt, text width=21.2em}{}
[{\parbox{6.2em}{\centering Physical State\\ $s_t^{\mathrm{phy}}$}}
    [{Entity and Attribute}
        [{Objects}
            [{Intrinsic properties}
                [{\taxleaf{name; size; weight; color; material}{e.g., cup; small; light; white; ceramic}}]
            ]
            [{Contextual attributes}
                [{\taxleaf{motion state; position; physical condition; semantic content}{e.g., static; on the table; cup broken; display says ``Next train in 5 minutes''}}]
            ]
        ]
        [{Characters}
            [{Intrinsic properties}
                [{\taxleaf{height; weight; clothing type and color}{e.g., about 1.75m; about 70kg; blue jacket}}]
            ]
            [{Contextual attributes}
                [{\taxleaf{facial expression; pose; gesture; gaze; motion state; position; body condition; semantic content}{e.g., frowning; standing; pointing to the door; looking at person 2; walking; near the door; injured; speech: ``Leave now''}}]
            ]
        ]
    ]
    [{Relations}
        [{Spatial relations}
            [{\taxleaf{relative position; distance; occlusion}{e.g., cup left of laptop; person 1 close to person 2; person 1 blocks the door}}]
        ]
        [{Contact relations}
            [{\taxleaf{contact relation}{e.g., hand touches the cup}}]
        ]
    ]
    [{Environment}
        [{Time / place / region}
            [{\taxleaf{time; place; region}{e.g., daytime; classroom; near the window}}]
        ]
        [{Ambient condition}
            [{Visual condition}
                [{\taxleaf{lighting; visibility}{e.g., bright; clear}}]
            ]
            [{Acoustic condition}
                [{\taxleaf{noise level; background sound}{e.g., noisy; traffic sound}}]
            ]
            [{Air condition}
                [{\taxleaf{temperature; wind; humidity; smoke}{e.g., hot; strong wind; humid; smoke present}}]
            ]
            [{Space condition}
                [{\taxleaf{crowdedness; clutter; openness}{e.g., crowded; cluttered; open space}}]
            ]
        ]
    ]
]
\end{forest}}
\caption{\textbf{Taxonomy of the physical world state $s_t^{\mathrm{phy}}$.} Entities (objects and characters) carry intrinsic and contextual attributes; relations record spatial and contact structure; the environment records time, place, region, and ambient visual, acoustic, air, and space conditions. Leaf boxes list the annotation fields (bold) with representative values (italics).}
\label{fig:physical-state-taxonomy}
\end{figure}

\begin{figure}[!t]
\centering
\scriptsize
\resizebox{\linewidth}{!}{%
\begin{forest}
forked edges,
for tree={
    grow=east,
    reversed=true,
    anchor=base west,
    parent anchor=east,
    child anchor=west,
    base=center,
    font=\scriptsize,
    rectangle,
    rounded corners,
    align=left,
    edge+={mwmvioletframe!50, line width=0.55pt},
    inner xsep=3pt,
    inner ysep=2.5pt,
    s sep=5pt,
    l sep=7pt,
    line width=0.55pt,
},
where level=0{fill=mwmvioletframe, draw=mwmvioletframe, text=white, font=\bfseries\scriptsize, text width=6.8em}{},
where level=1{fill=mwmviolet, draw=mwmvioletframe, font=\bfseries\scriptsize, text width=12.2em}{},
where level=2{fill=mwmviolet!45, draw=mwmvioletframe!65, text width=9.2em}{},
where level=3{fill=gray!7, draw=gray!55, text width=9.8em}{},
where n children=0{fill=white, draw=gray!50, inner xsep=4pt, text width=21.2em}{}
[{\parbox{6.2em}{\centering Mental State\\ $s_t^{\mathrm{ment}}$}}
    [{Mental Entity and Attribute}
        [{Individual $m_t^i$}
            [{Identity attributes}
                [{\taxleaf{name; occupation}{e.g., Alice; teacher}}]
            ]
            [{Epistemic state}
                [{\taxleaf{beliefs; attention focus}{e.g., believes the train is delayed; looking at the display}}]
            ]
            [{Motivational state}
                [{\taxleaf{goals; intentions}{e.g., wants to catch the train; plans to wait}}]
            ]
            [{Affective state}
                [{\taxleaf{emotions}{e.g., angry; anxious; relieved}}]
            ]
            [{Dispositional state}
                [{\taxleaf{preferences; values; personality}{e.g., prefers quiet places; values fairness; competitive}}]
            ]
            [{Normative state}
                [{\taxleaf{rules; cultural norms; customs}{e.g., basketball rules; greet others politely; remove shoes indoors}}]
            ]
            [{Behavioral constraints}
                [{\taxleaf{obligations; prohibitions}{e.g., must follow the rules; cannot push others}}]
            ]
        ]
        [{Group $m_t^G$}
            [{\taxleaf{same field structure as individuals, held by a collective actor}{e.g., Team A (basketball team) believes it is losing; wants to win; excited; values teamwork; follows tournament rules; must stay in position}}]
        ]
    ]
    [{Relations}
        [{Attitudes}
            [{\taxleaf{person-object; person-person; person-group; group-group attitude}{e.g., person 1 likes the gift; person 1 distrusts person 2; person 1 supports Team A; Team A dislikes Team B}}]
        ]
        [{Role relations}
            [{\taxleaf{person-person; person-group; group-group role relation}{e.g., teacher-student; person 1 is coach of Team A; Team A and Team B are opponents}}]
        ]
    ]
    [{Atmosphere}
        [{\taxleaf{scene-level social atmosphere}{e.g., tense; cooperative; festive; awkward}}]
    ]
]
\end{forest}}
\caption{\textbf{Taxonomy of the mental world state $s_t^{\mathrm{ment}}$.} Individual mental entities carry identity, epistemic, motivational, affective, dispositional, normative, and constraint fields; group entities reuse the same structure at the collective level; mental relations record attitudes and role relations; a scene-level atmosphere summarizes collective mood. Leaf boxes list the annotation fields (bold) with representative values (italics).}
\label{fig:mental-state-taxonomy}
\end{figure}
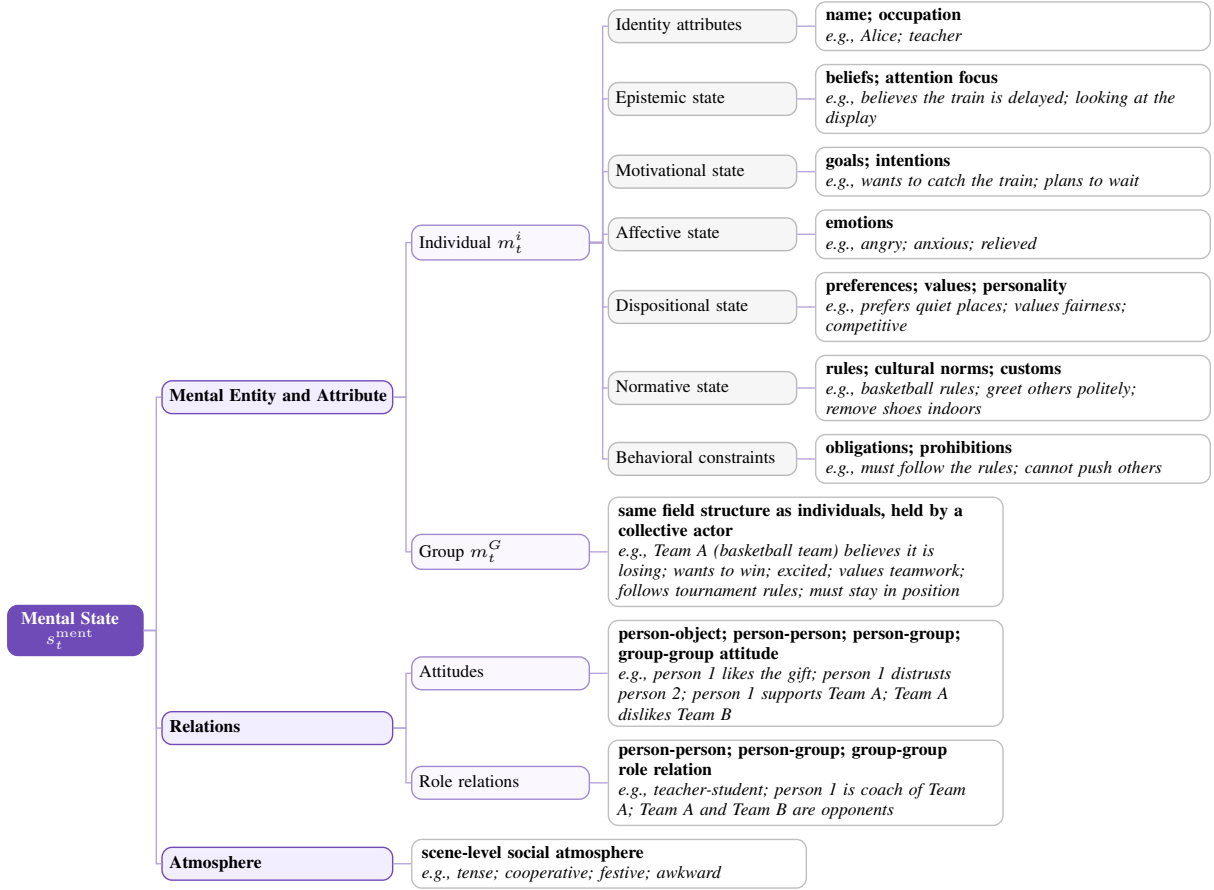

\paragraph{Core interactive variables.}
\label{sec:state-taxonomy}
At time $t$, the global world state is a product state
\begin{equation}
s_t =
\big(s_t^{\mathrm{phy}},s_t^{\mathrm{ment}}\big)
\in
\mathcal{S}^{\mathrm{phy}}\times\mathcal{S}^{\mathrm{ment}}.
\label{eq:joint-state}
\end{equation}
The physical component records what exists, where it is, how it is configured, and what perceptible signals it carries:
\begin{equation}
s_t^{\mathrm{phy}}
=
\big(
O_t,C_t,R_t^{\mathrm{phy}},E_t
\big),
\label{eq:physical-state}
\end{equation}
where $O_t$ denotes objects, $C_t$ denotes characters or embodied agents, $R_t^{\mathrm{phy}}$ denotes spatial and contact relations, and $E_t$ denotes environmental conditions. Figure~\ref{fig:physical-state-taxonomy} instantiates this physical component as a taxonomy tree, making explicit how entity attributes, relations, and environmental conditions branch into concrete annotation fields and representative values.

The mental component records latent variables that explain how agents interpret the physical scene and how they are disposed to act:
\begin{equation}
s_t^{\mathrm{ment}}
=
\big(
\{m_t^i\}_{i\in\mathcal{I}},
\{m_t^G\}_{G\in\mathcal{G}},
R_t^{\mathrm{ment}},
\alpha_t
\big),
\label{eq:mental-state}
\end{equation}
where $m_t^i$ is an individual-level mental state, $m_t^G$ is an optional group-level mental state, $R_t^{\mathrm{ment}}$ stores social attitudes and role relations, and $\alpha_t$ denotes the atmosphere of the environment. An individual mental state can be written as
\begin{equation}
m_t^i =
\big(
\mathrm{id}_t^i,
b_t^i,
q_t^i,
g_t^i,
\iota_t^i,
e_t^i,
d_t^i,
n_t^i,
c_t^i
\big),
\label{eq:individual-mental-state}
\end{equation}
where $\mathrm{id}$ denotes identity attributes, $b$ beliefs, $q$ attention focus, $g$ goals, $\iota$ intentions, $e$ emotions, $d$ dispositions, $n$ norms, and $c$ behavioral constraints. Figure~\ref{fig:mental-state-taxonomy} gives the corresponding mental-state taxonomy: individual and group mental entities carry identity, epistemic, motivational, affective, dispositional, normative, and constraint fields, while relation and atmosphere branches encode social attitudes, roles, and scene-level mood. The mental state is not directly visible. It is a compact state variable that summarizes the history relevant to future interpretation and action, such as a person's prior belief that a train is delayed or a team's belief that it is losing.

The target observation is a first-person rendering of the global state:
\begin{equation}
o_t^{\target}
=
\big(o_t^{\target,\mathrm{phy}},o_t^{\target,\mathrm{ment}}\big)
\in
\mathcal{O}^{\mathrm{phy}}\times\mathcal{O}^{\mathrm{ment}}.
\label{eq:observation-variable}
\end{equation}
The physical observation $o_t^{\target,\mathrm{phy}}$ contains perceptible signals such as visual layout, speech, sound, gesture, gaze, and written text. The mental observation $o_t^{\target,\mathrm{ment}}$ contains the target's introspection and mental inferences:
\begin{equation}
o_t^{\target,\mathrm{ment}}
=
\Big(
\underbrace{\omega_{\mathrm{self}}(m_t^{\target})}_{\text{self-observation}},
\underbrace{\{\widehat{m}_{t,\target}^{j,(\ell)}\}_{j\neq\target,\ 1\leq \ell\leq L}}_{\text{Theory-of-Mind inferences}}
\Big).
\label{eq:mental-observation-variable}
\end{equation}
Here $\widehat{m}_{t,\target}^{j,(1)}$ denotes what the target infers about agent $j$; $\widehat{m}_{t,\target}^{j,(2)}$ can denote what the target thinks agent $j$ thinks about the target or another agent. These inferred variables need not equal the true $m_t^j$. A target may see Bob crying and infer sadness, while another observer infers relief, because each observer has different access to context, relations, and prior beliefs.

Finally, the target action is represented as a coupled pair
\begin{equation}
a_t^{\target}
=
\big(a_t^{\target,\mathrm{phy}},a_t^{\target,\mathrm{ment}}\big)
\in
\mathcal{A}^{\mathrm{phy}}\times\mathcal{A}^{\mathrm{ment}}.
\label{eq:action-variable}
\end{equation}
The two components are not independent actions added together; they are two dimensions of the same action. The physical component is the carrier, i.e., moving, pushing, grasping, looking, gesturing, speaking, typing, or changing facial expression. The mental component is the semantic, intentional, affective, or cognitive content carried by that physical act: requesting, comforting, deceiving, threatening, apologizing, focusing attention, or expressing approval. A mental action without a physical carrier cannot directly move an object or change air pressure; it can affect others only when expressed through a perceptible carrier or internalized as the target's own cognitive control.

\subsubsection{Dynamic Joint Modeling}
\label{sec:mwm-joint-modeling}

The MWM process at each time step consists of three coupled functions: observation generation, action proposal, and state transition. Together they define a target-centric partial-observation simulator:
\begin{equation}
s_t
\xrightarrow{\Omega_{\target}}
o_t^{\target}
\xrightarrow{\Pi_{\target}}
a_t^{\target}
\xrightarrow{T_{\theta}}
s_{t+1}.
\label{eq:mwm-process}
\end{equation}

\paragraph{Observation generation function.}
The world model renders the target's observation from the current joint state:
\begin{equation}
o_t^{\target}
=
\Omega_{\target}(s_t)
=
\left(
\Omega_{\target}^{\mathrm{phy}}(s_t^{\mathrm{phy}};\kappa_t^{\target}),
\Omega_{\target}^{\mathrm{ment}}(s_t^{\mathrm{phy}},s_t^{\mathrm{ment}};\rho_t^{\target})
\right),
\label{eq:observation}
\end{equation}
where $\kappa_t^{\target}$ summarizes the target's perceptual access, such as position, line of sight, hearing range, and modality availability, and $\rho_t^{\target}$ summarizes the target's social and cognitive perspective, such as role, prior belief, relationship to others, norm awareness, and ToM capacity.

The physical observation function $\Omega_{\target}^{\mathrm{phy}}$ maps the physical world to signals the target can directly perceive. It depends on the physical state and the target's access conditions. When Bob lowers his head and speaks quietly, the lowered head and sound pattern are physical observations. Whether Bob is ashamed, relieved, or deceptive is not itself a physical observation.

The mental observation function $\Omega_{\target}^{\mathrm{ment}}$ produces a target-specific mental readout. Its dependence on $s_t^{\mathrm{ment}}$ does not mean that the target reads global mental states directly. Rather, the world model uses the global mental state to generate what this target can infer under partial observability:
\begin{equation}
\widehat{m}_{t,\target}^{j,(1)}
\sim
Q_{\target}^{(1)}
\left(
m_t^j
\mid
o_t^{\target,\mathrm{phy}},
m_t^{\target},
R_t^{\mathrm{ment}},
\alpha_t
\right).
\label{eq:first-order-tom}
\end{equation}
This equation is intentionally perspectival. If Alice knows that Jack expected a prize and sees him crying, she may infer disappointment; if Bob knows Jack has just been saved from danger, he may infer relief. The same physical signal can yield different mental observations because the interpretation is mediated by beliefs, goals, roles, and scene context.

\paragraph{Action proposal function.}
The target agent generates a single-step action from its observation; systems that enumerate candidate actions rather than sample a single one use the set-valued form:
\begin{equation}
a_t^{\target}
\sim
\Pi_{\target}\left(\cdot\mid o_t^{\target}\right),
\qquad
\mathcal{A}_t^{\target}
=
\Pi_{\target}^{K}(o_t^{\target})
=
\left\{a_{t,1}^{\target},\ldots,a_{t,K}^{\target}\right\}.
\label{eq:target-policy}
\end{equation}
The conditioning variable is the observation, not the full state. This is the formal reason MWM can represent false belief, ignorance, surprise, and socially biased action. The target may choose a poor action because its $o_t^{\target}$ omits an object, misreads another person's intention, or contains a mistaken belief.

Because the physical and mental components are two dimensions of one action, the policy in Eq.~\eqref{eq:target-policy} is a \emph{joint} distribution over the pair: both dimensions are determined together, conditioned on the same observation, which already carries the target's goals, beliefs, and inferred mental context. When the joint policy needs to be decomposed, the chain rule gives the exact identity
\begin{equation}
\Pi_{\target}\big(a_t^{\target,\mathrm{phy}},a_t^{\target,\mathrm{ment}}\mid o_t^{\target}\big)
=
\Pi_{\target}^{\mathrm{phy}}
\left(a_t^{\target,\mathrm{phy}}\mid o_t^{\target}\right)
\Pi_{\target}^{\mathrm{ment}}
\left(a_t^{\target,\mathrm{ment}}\mid a_t^{\target,\mathrm{phy}},o_t^{\target}\right),
\label{eq:action-factorization}
\end{equation}
and the symmetric factorization that conditions the carrier on the content is equally valid; neither ordering claims that the target generates one component before the other. We display the carrier-conditioned form because it matches how actions are \emph{consumed} in our setting: an observed or candidate behavior exposes its physical carrier, and the mental content must be read off conditioned on that carrier, which is exactly the action-decomposition step of Section~\ref{sec:mentis}. The conditional is genuinely informative in both directions: saying ``leave now'' and saying the same words as a joke share a similar physical speech carrier but differ in mental action, while comforting may be realized through speech, a gesture, or proximity, so the same mental action can be carried by different physical actions.

\paragraph{State transition function.}
The world model predicts the next joint state from the current joint state and the target action, $s_{t+1}\sim T_{\theta}(s_t,a_t^{\target})$. Writing $T_{\theta}(s_{t+1}\mid s_t,a_t^{\target})$ for the conditional density of this kernel, we use a staged factorization because it makes the physical and mental causal channels inspectable:
\begin{align}
T_{\theta}(s_{t+1}\mid s_t,a_t^{\target})
&=
T_{\theta}^{\mathrm{phy}}
\left(
s_{t+1}^{\mathrm{phy}}
\mid
s_t^{\mathrm{phy}},
s_t^{\mathrm{ment}},
a_t^{\target,\mathrm{phy}}
\right)
\nonumber\\
&\quad\cdot
T_{\theta}^{\mathrm{ment}}
\left(
s_{t+1}^{\mathrm{ment}}
\mid
s_t^{\mathrm{phy}},
s_t^{\mathrm{ment}},
a_t^{\target,\mathrm{phy}},
a_t^{\target,\mathrm{ment}}
\right).
\label{eq:joint-transition}
\end{align}
The physical transition is conditioned on $s_t^{\mathrm{phy}}$ because objects, space, visibility, force, and affordance constrain what can physically happen. It is also conditioned on $s_t^{\mathrm{ment}}$ because other agents' latent goals and intentions may produce concurrent reactions that affect the next physical state. It is not directly conditioned on $a_t^{\target,\mathrm{ment}}$: the semantic intention ``comfort Bob'' does not itself move air, change posture, or reposition objects unless it is realized through speech, gesture, motion, or another physical carrier already represented by $a_t^{\target,\mathrm{phy}}$.

The mental transition is conditioned on both action components. Current physical state supplies the stimulus and situational grounding; current mental state supplies continuity of belief, affect, intention, norm, and relationship; the physical action supplies the observable behavior; and the mental action supplies the intended meaning carried by that behavior. If Alice says ``Do not worry, I am here'' while approaching Bob in a dark room, the sound and approach are physical carriers, while reassurance and support are mental-action content. Bob's next mental state may shift from fear and isolation toward partial relief only because these channels are interpreted together.

Although Eq.~\eqref{eq:joint-transition} is written around the target action, it does not assume that other agents are passive. A more explicit multi-agent transition would condition on actions $a_t^{-\target}$ from all non-target agents:
\begin{equation}
T_{\theta}(s_{t+1}\mid s_t,a_t^{\target})
=
\int
T_{\theta}(s_{t+1}\mid s_t,a_t^{\target},a_t^{-\target})\,
p(a_t^{-\target}\mid s_t)\,
\mathrm{d}a_t^{-\target},
\label{eq:target-centric-marginalization}
\end{equation}
where $p(a_t^{-\target}\mid s_t)$ denotes the simulator's predictive distribution over the other agents' actions. Conditioning on $s_t$ does not grant those agents access to the global state, and thus creates no conflict with partial observability: each non-target agent $j$ still acts only from its own partial rendering, and the distribution is the composition of observation rendering and policy, $p(a_t^{-\target}\mid s_t)=\prod_{j\neq\target}\Pi_{j}\big(a_t^{j}\mid \Omega_{j}(s_t)\big)$ for simultaneous moves, with $\Omega_j$ the rendering of Eq.~\eqref{eq:observation} applied to agent $j$. The world model can evaluate this composition precisely because it is third-person and omniscient; the formulation remains target-centric because the other agents' observations and policies stay inside the marginal. MWM uses this form when the task asks what the specified target will do or what happens after the target's candidate action. The likely reactions of others are marginalized through their mental states: their goals, intentions, obligations, and attitudes are part of $s_t^{\mathrm{ment}}$ and can influence $s_{t+1}$.

\section{Where Mental World Modeling Matters}
\label{sec:applications}

The application value of MWM should not be measured by whether one can name many social domains. Almost every human-facing system is ``social'' in a loose sense. The stronger claim is narrower and more useful: \emph{MWM matters when the quality of an action depends on variables that are not recoverable from the physical scene alone, yet still determine what a human will perceive, choose, accept, resist, or learn}. In decision-theoretic terms, the value of mental-state information is the expected utility gap between the best intervention chosen with access to the coupled physical-mental state and the best intervention chosen from the physical state alone. This gap is large exactly when beliefs, goals, attention, trust, obligations, affect, or norms change which action is useful, and it is small when the next action is determined almost entirely by physical feasibility. A grasping controller does not need MWM to close its fingers around a cup; a household assistant needs it to decide whether handing over that cup is helpful, intrusive, rude, unsafe, or misleading.

\begin{mwminsightbox}[When MWM is worth the cost]
MWM is most valuable when three conditions hold together: \textbf{partial observability} creates a gap between the global state and the target's perceived state; \textbf{mental variables} change which action is rational or acceptable; and \textbf{actions carry social meaning} beyond their physical effects. If any of these conditions is absent, a simpler physical or task-specific world model may be sufficient.
\end{mwminsightbox}

The remainder of this section argues that these conditions arise, in different combinations, across the settings where AI systems are already deployed. For each regime, we identify what a physical-only world model computes, where it becomes under-specified, and which MWM variables restore the missing decision-relevant structure.

\paragraph{Embodied collaboration: joint progress is physical and mental.}
Physical world models already support control, planning, and generated interactive environments~\citep{ha2018world,hafner2023dreamerv3,bruce2024genie,deepmind2025genie3}. In human-facing embodiment, however, the transition to be modeled is not only ``what moves next''. A robot entering a kitchen, clinic room, or shared office must decide whether a person has noticed it, whether an object is socially available, whether an interruption is acceptable, and whether an action will be read as help or pressure. These distinctions are not cosmetic: a robot that silently moves a misplaced medicine bottle may be physically efficient but socially unsafe if the user does not know why the bottle moved. All three conditions hold: the user's attention and knowledge are partially observable, they change which assistance is appropriate, and the robot's own motion carries communicative meaning. MWM recasts these as state and transition questions: what does the target currently believe, what will the target observe, and how will the action revise both physical and mental states? Collaboration sharpens the point, because collaborative action requires a model of \emph{joint progress}: the environment changes, but so do attention, commitments, expectations, and shared knowledge. A physical-only planner can represent that a door is open; an MWM can represent that Alice opened it \emph{for} Bob, that Bob noticed the gesture, and that Bob now expects Alice to follow; such variables determine whether a failure should be repaired by moving an object, saying something, waiting, asking, apologizing, or handing control back to a human.

\paragraph{Care, support, and advising: the failure is rarely factual.}
MWM is not a replacement for domain expertise, clinical judgment, or professional responsibility; its value in care and assistive settings is representational. Many failures in advice-giving are not failures to retrieve facts but failures to model what the person understands, fears, assumes, or is socially pressured to do. A patient may hear the same instruction as reassurance, blame, coercion, or permission; an older adult may refuse a reminder not because it is physically inconvenient, but because it signals loss of autonomy. Here the mental variables (comprehension, trust, anxiety, obligation, dependency) are precisely the ones a physical record cannot expose, yet they decide whether an intervention helps or harms. An MWM-style system can maintain explicit hypotheses about these variables and expose them as auditable state rather than burying them inside a generic response. The strongest use is conservative: deciding when \emph{not} to act autonomously. If the model is uncertain about consent, distress, or role responsibility, the correct action may be to ask, defer, escalate, or present options rather than optimize a task metric. MWM thus supports a safer interface between automation and human authority, giving the system a place to represent uncertainty over human context and a mechanism for propagating that uncertainty into action choice.

\paragraph{Education and training: interventions target the learner's successor state.}
Tutoring is a natural MWM domain because the immediate physical state is often uninformative. The same wrong answer may reflect a missing concept, a brittle heuristic, low confidence, inattention, or frustration; these cases call for different interventions. A corrected derivation may help a confused learner, discourage an anxious one who already understands the concept, or be irrelevant to one attending to a different subproblem. MWM clarifies what an educational agent should simulate: an explanation is an action with a linguistic carrier and a \emph{mental transition target}. It should update knowledge, but it also updates confidence, motivation, trust in the tutor, and willingness to attempt the next step. The value of an intervention is therefore not only whether the next answer becomes correct, but whether the learner's successor state is better prepared for future learning. This makes MWM a natural formal language for adaptive explanation, formative feedback, role-play training, and simulation-based assessment.

\paragraph{Interactive agents and social worlds: persistence requires mental transition.}
Large generative models increasingly power games, digital humans, and interactive social environments, including agent societies that simulate believable daily behavior~\citep{openai2024sora,bruce2024genie,deepmind2025genie3,worldlabs2025marble,park2023generative}. These systems need more than visual continuity. A digital character should remember promises, hide information it has not observed, react differently to friends and strangers, and maintain coherent emotional and normative trajectories. Prompt-level personality descriptions create local style, but they do not define a transition system. MWM provides that transition interface: it separates the scene from each character's observation, decomposes action into physical carrier and mental-social content, and updates both successor states. This yields controllability, since designers can inspect why a branch produced suspicion rather than relief, why a character refused an offer, or why a group atmosphere shifted after a public insult; it also connects interactive generation to social evaluation. A world is not convincing merely because frames look realistic; it is convincing when agents' beliefs, relationships, and norms evolve in ways that remain legible across counterfactual branches.

\section{\Mentis: An Inspectable Baseline Implementation of MWM}
\label{sec:mentis}

In this section, we propose \Mentis{}, a modular, inspectable, and training-free baseline that implements the theoretical process described above. \Mentis{} is designed to be a practical testbed for evaluating whether existing LLMs and MLLMs can be constrained to behave as a MWM. 
It follows the theoretical framework by parsing a situated scene into a joint physical-mental state, rendering the target agent's partial observation, simulating action-conditioned futures, and selecting an answer based on these simulations.

\begin{figure}[!t]
    \centering
    \includegraphics[width=\linewidth]{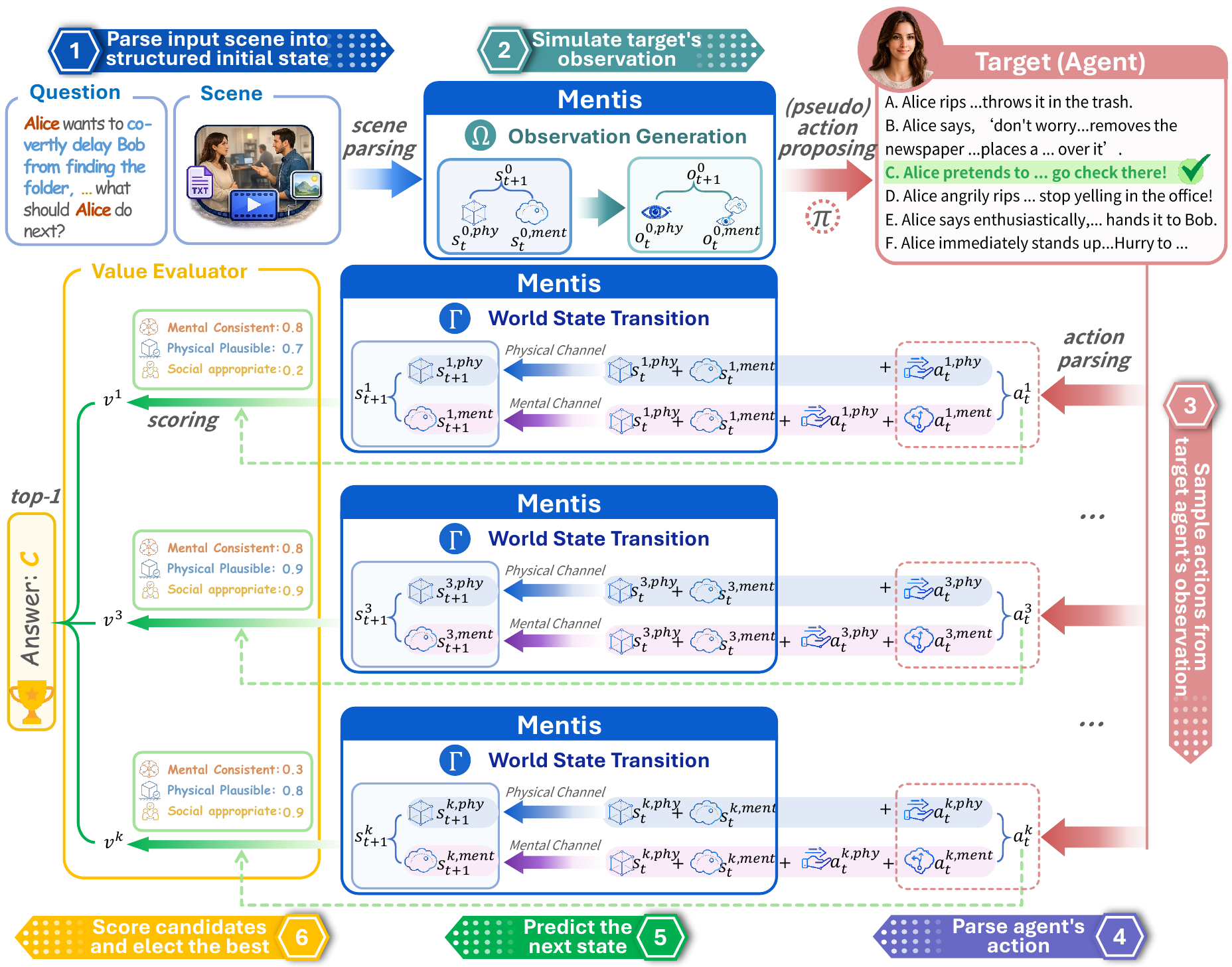}
    \caption{\textbf{\Mentis{} pipeline.} The system converts an input scene into a structured current state $\hat{s}_t$, renders the target pseudo-agent's observation $\hat{o}_t^{\target}$, parses each option into an action branch, simulates physical and mental successor states in parallel, and evaluates the resulting futures to select the final action. The numbered regions correspond to the six stages described in Section~\ref{sec:mentis-pipeline}.
    }
    \label{fig:mentis}
\end{figure}

\subsection{\Mentis{} Design Principle}
\label{sec:mentis-design}

A direct-answer baseline receives a scene, question, and options, and outputs an option label. \Mentis{} instead exposes the latent reasoning path, following the formal process as in Section~\ref{sec:theory}:
\[
x
\rightarrow
\hat{s}_t
\rightarrow
\hat{o}_t^{\target}
\rightarrow
\{\hat{a}_{t,k}^{\target}\}_{k=1}^{K}
\rightarrow
\{\hat{s}_{t+1}^{k}\}_{k=1}^{K}
\rightarrow
\{r_k\}_{k=1}^{K}
\rightarrow
\hat{k},
\]
where $x$ denotes the input sample, $\target$ the target agent, $K$ the number of candidate options, $r_k$ the branch score that the value evaluator assigns to candidate $k$ from its simulated consequence (Section~\ref{sec:mentis-value-evaluator}), and $\hat{k}$ the selected branch. The hats emphasize that every intermediate object is a model prediction, not ground truth.

\begin{quote}
\itshape
\Mentis{} should never select an answer before constructing the branch it is selecting. The decision must be grounded in an explicit path $s_t \rightarrow o_t^{\target} \rightarrow a_{t,k}^{\target} \rightarrow s_{t+1}^{k}$.
\end{quote}

This principle makes \Mentis{} useful even where its absolute accuracy is modest. If a prediction fails, the failure can be localized: the scene may have been parsed into the wrong current state; the target observation may have leaked global information; the option may have been decomposed into the wrong physical or mental action; the physical or mental transition may be implausible; or the evaluator may have scored the simulated branches incorrectly. A monolithic answer prompt offers no such decomposition.

\subsection{Pipeline Overview}
\label{sec:mentis-pipeline}

An input record contains a scene medium (text, images, or a sounding video), a target agent, a question, and a finite set of $K$ natural-language options that together define the candidate action space. Inference then follows six stages, matching the numbered regions of Figure~\ref{fig:mentis}; each stage instantiates, with model predictions, the corresponding operator of the formal process in Section~\ref{sec:math-formalization}.

\begin{enumerate}[label=\arabic*), leftmargin=1.6em, itemsep=1pt, topsep=2pt]
    \item \textbf{State parsing.} The \textsc{StateParser} maps the raw scene into the joint current state $\hat{s}_t=(\hat{s}_t^{\mathrm{phy}},\hat{s}_t^{\mathrm{ment}})$, following the taxonomies of Figures~\ref{fig:physical-state-taxonomy} and~\ref{fig:mental-state-taxonomy}.
    \item \textbf{Observation generation.} The \textsc{ObservationGenerator} within MWM renders the target's first-person observation $\hat{o}_t^{\target}$ from $\hat{s}_t$, filtering out information the target cannot access.
    \item \textbf{Action decomposition.} The target pseudo-agent and \textsc{ActionParser} convert each option $c_k$ into a structured candidate action $\hat{a}_{t,k}^{\target}$ with explicit physical and mental components.
    \item \textbf{Branch simulation.} For every candidate action, the \textsc{WorldStateTransitor} modules within MWM predict the successor state: the physical and mental channels are each predicted from the shared time-$t$ inputs of Figure~\ref{fig:mentis}, and the two components are merged into $\hat{s}_{t+1}^{k}$. The $K$ branches run in parallel.
    \item \textbf{Value evaluation.} The \textsc{Evaluator} scores each imagined future along three normalized criteria and emits a safety veto flag (Section~\ref{sec:mentis-value-evaluator}).
    \item \textbf{Decision.} A deterministic decision module selects the highest-value branch under fixed tie-breaking rules and maps it back to the option label.
\end{enumerate}

Every stage writes its artifact (parsed state, rendered observation, decomposed actions, successor states, score table, and decision trace) to the run directory, which is what enables the component-level evaluation and ablations.

\subsection{Mental World Model Core}
\label{sec:mentis-world-model-core}

\paragraph{State parsing as structured scene understanding.}
The state parser is the perceptual entry point of \Mentis. It does not merely summarize the input story; it converts the raw scene into the state schema of Section~\ref{sec:math-formalization}: a physical component containing entities, attributes, relations, and environmental conditions, and a mental component containing agent-level and scene-level variables such as beliefs, attention, goals, intentions, emotions, preferences, relations, norms, and atmosphere. The current implementation represents this state as JSON because JSON is easy to validate, diff, log, and compare against annotations; the deeper commitment is not JSON itself but the requirement that the state be explicit enough for later modules to read and update. Parsing is modality-aware: textual stories are parsed directly; image stories combine visual evidence with caption narration; videos are decoded into adaptively sampled key frames plus a transcribed audio track, so that dialogue and ambient sound enter the state (details in Appendix~\ref{app:mentis-extended}). This stage is intentionally difficult: some mental variables are directly supported by text, speech, facial expression, or gaze, while others are only weakly implied. \Mentis{} therefore treats $\hat{s}_t^{\mathrm{ment}}$ as a hypothesis-bearing state, and parser errors are analyzed as a first-class failure mode rather than hidden behind final-answer accuracy.

\paragraph{Observation generation as information filtering.}
The observation generator operationalizes partial observability. It receives the third-person state $\hat{s}_t$ and emits what the target can see, hear, know, remember, or infer. The physical part is a target-specific projection of the physical state: visible objects, positions, gestures, utterances, and motion. The mental part is \emph{not} the global mental state; it is the target's self-state plus perspective-limited inferences about others. A correct observation may therefore omit facts that are true in $\hat{s}_t$ but inaccessible to the target, and may contain false or incomplete beliefs when the target is misinformed. This module marks a crucial difference between MWM and direct social reasoning: if the target lacks a fact that the dataset narrator provides, a direct-answer model can still exploit that fact, whereas \Mentis{} makes the access relation explicit. The target's candidate actions are grounded in its rendered observation, while the world model retains the richer global state needed to simulate each branch's consequences.

\paragraph{Action decomposition as a physical-mental interface.}
Each natural-language option is parsed into a candidate action with a physical component (the carrier: moving, speaking, handing, pointing, opening, hiding, waiting) and a mental component (the meaning or intended cognitive-social effect: reassuring, deceiving, refusing, apologizing, drawing attention, maintaining face). The decomposition matters because the two components have different causal roles in the transition: the physical transition is constrained primarily by the carrier, whereas the mental transition depends on both the carrier and the meaning that other agents may attribute to it.

\paragraph{Coupled state transition.}
The implementation separates the transition into physical and mental submodules but does not assume the two dynamics are independent. Consistent with Eq.~\eqref{eq:joint-transition} and Figure~\ref{fig:mentis}, both submodules condition on the full joint state at time $t$: the physical submodule predicts $\hat{s}_{t+1}^{\mathrm{phy}}$ from $(\hat{s}_t^{\mathrm{phy}},\hat{s}_t^{\mathrm{ment}},\hat{a}_t^{\target,\mathrm{phy}})$, because other agents' latent goals and intentions shape what physically happens next, and the mental submodule predicts $\hat{s}_{t+1}^{\mathrm{ment}}$ from $(\hat{s}_t^{\mathrm{phy}},\hat{s}_t^{\mathrm{ment}},\hat{a}_t^{\target,\mathrm{phy}},\hat{a}_t^{\target,\mathrm{ment}})$, because interpretation depends on both the carrier and its meaning. The coupling thus lives in the shared cross-channel conditioning, not in feeding one submodule's output to the other; a merger composes the two predictions into the joint successor $\hat{s}_{t+1}$. If Alice says ``Don't worry'' while moving a newspaper over a folder, the physical channel describes the newspaper covering the folder and the utterance becoming audible; the mental channel asks how Bob interprets the act (reassured, suspicious, or unaware) given the same scene, the prior mental context, and both aspects of Alice's action.

\subsection{Target Pseudo-Agent}
\label{sec:mentis-target-agent}

The target module is called a \emph{pseudo}-agent because, in our implementation and the benchmark setting, it does not freely invent actions: it receives the option set supplied by the evaluation sample and turns those options into candidate branches. This choice is methodological. For multiple-choice situated reasoning, the goal is not to measure whether a model can generate a creative policy, but to compare given branches under a controlled action space, which keeps final accuracy comparable across runs and makes every branch available for transition-level analysis. Conceptually, \Mentis{} also supports a free-proposal regime in which the pseudo-agent samples $K$ candidate actions from its own observation, which is the natural extension for interactive agents and open-world planning. Keeping the two regimes separate avoids a common evaluation error: confusing the quality of action generation with the quality of world-state transition.

\subsection{Value Evaluator and Decision Rule}
\label{sec:mentis-value-evaluator}

The value evaluator is never asked to choose the best option directly from the question. It receives, for each branch, a package containing the current state, the target observation, the decomposed action, the simulated successor state, and the question, and assigns three normalized scores: \emph{mental consistency} (would the target, given its beliefs and goals, choose this action, and does the mental successor state follow?), \emph{physical plausibility} (is the simulated outcome materially possible?), and \emph{social appropriateness} (does the action fit the scene's social relations, politeness norms, customs, moral constraints, role obligations, and task context?). These dimensions deliberately mirror the theory: a branch can be physically plausible but mentally inconsistent (an action the target could perform but would not choose given its belief), or mentally consistent but socially inappropriate (an action explained by the character's anger that nevertheless violates a role obligation or moral constraint). The evaluator additionally emits a binary safety/legality veto that zeroes a branch's value regardless of its other scores.
Scoring is comparative: the evaluator sees all successful branches in one shared context and produces per-dimension grades together with a strict cross-branch ranking. The final value of a branch is a weighted combination of the three dimension scores after the veto, and the decision module is deterministic and kept outside the language model, so the selection is reproducible and the score table remains auditable. Fallback behavior, tie-breaking rules, per-run logging, and further implementation details are given in Appendix~\ref{app:mentis-extended}.

\section{Evaluation Settings}
\label{sec:benchmark}

This section shows the experimental settings of Section~\ref{sec:experiments}, including the evaluation data, the task protocol and metric suite, the systems under comparison, and the operating point of every reported run. 

\subsection{Menti-Bench: A Validation Testbed Dataset}
\label{sec:dataset}

Existing outcome-only social-reasoning benchmarks are insufficient for evaluating MWM: a system may select the correct option from surface cues without constructing the variables the framework claims are necessary. We therefore assemble \textbf{Menti-Bench}, a small \emph{process-complete} testbed for target-centric world modeling. Each record contains a situated story in exactly one of three modalities (a textual narrative; a sequence of one to five images with a short scene anchor; or a sounding video clip whose dialogue and ambient audio carry part of the evidence), a designated target agent, a deliberately minimal decision question, and six candidate actions written in natural language. The gold annotation covers the full MWM process (the joint current state $s_t^{\star}$, the target observation $o_t^{\target,\star}$, one successor state per option, and the final action), follows the state taxonomies of Section~\ref{sec:math-formalization}, and is reserved for evaluation and oracle interventions; a standard run receives only the public scene, target, question, and options.

The testbed comprises 448 records (320 text, 100 image, 28 sounding video), each carrying the full process-level gold annotation (2{,}688 annotated successor states in total). Records are stratified over four scene categories (interpersonal, object/resource, spatial/perceptual, risk/norm) and five everyday domains; 78\% of scenes involve at least two characters, so most records require mental inference about other agents. Construction prioritized shortcut control over dataset size: gold letters are near-uniform over the six positions, gold and distractor option lengths are matched, textual records passed adversarial option balancing and blind options-only probes, and media records passed per-item human quality checks (only 28 of 50 produced sounding videos were retained).
The construction protocol, per-modality quality control, composition figure, and full summary statistics are given in Appendix~\ref{app:benchmark-details}; a datasheet is given in Appendix~\ref{app:datasheet}.

\subsection{Task Protocol and Metric Suite}
\label{sec:metrics}
\label{sec:benchmark-task}

\paragraph{Task protocol.}
All systems face one constrained task: given the story, the target agent, the question, and the six options, select the option describing the target's most likely next action. A direct-answer baseline may output only the chosen letter. An MWM system must additionally expose its full trace (the parsed current state, the target observation, the six decomposed actions, the six simulated successor states, optional branch scores, and the final choice) as machine-checkable artifacts, which makes the intermediate objects of Section~\ref{sec:math-formalization} available for evaluation whenever a system claims to construct them. The evaluation objects and their output fields are itemized in Appendix~\ref{app:benchmark-details}.

\paragraph{Metric suite.}
The metrics are layered, because a system can be correct at the final answer while failing the MWM process, and can construct plausible states while choosing the wrong branch. \emph{Outcome}: final-action F1 over the six option letters, reported with per-slice breakdowns; run-failure rate is reported separately rather than folded into the score; every headline comparison is paired, system versus system on identical records. \emph{Structure}: deterministic checks that the claimed artifacts exist and are well-formed, including per-option branch schema coverage. \emph{Semantics}: calibrated LLM judges~\citep{zheng2023judging} compare predicted states, observations, and successor states against the gold process annotations, covering physical and mental fidelity, semantic consistency, transition reasonableness, and physical-mental coupling; judge prompts encode the benchmark's annotation conventions, and judge scores are validated against human ratings before entering any headline claim. \emph{Valuation}: branch-score alignment with reference judgments, the decision margin between the chosen branch and the runner-up, and the tie rate the deterministic decision rule must resolve. \emph{Validity audits}: an options-only floor (S0) measures how well records can be answered from the options alone, and channel interventions (image-to-caption replacement; audio, frame-order, and visual-stream removal) measure how much performance depends on the intended evidence channels. Exact definitions and computation are given in Appendix~\ref{app:metrics}, and probe protocols in Appendix~\ref{app:probe-details}.

\subsection{Systems, Models, and Operating Point}
\label{sec:exp-setup}

\paragraph{Systems.}
Table~\ref{tab:exp-systems} lists the evaluated systems as a structural ladder. Each rung adds one modeling commitment, so adjacent comparisons attribute performance to a specific mechanism rather than to prompting in general; rungs S2 and S3 instantiate chain-of-thought prompting~\citep{wei2022chain} and self-consistency~\citep{wang2023selfconsistency}, so the ladder embeds the standard test-time reasoning baselines. Information-removal ablations (A1--A3) and oracle substitutions (O1--O4) are interventions on the full system.

\begin{table}[!t]
\caption{System ladder and interventions. Each rung adds one commitment; ablations and oracles intervene on the full system (S6).}
\label{tab:exp-systems}
\centering
\small
\begingroup
\rowcolors{2}{gray!6}{white}
\begin{tabularx}{0.98\linewidth}{llX}
\toprule
\textbf{ID} & \textbf{System} & \textbf{What it adds or removes} \\
\midrule
S0 & Options-only floor & No story at all; measures guessability of the option set. \\
S1 & Direct answer & Story + question + options; single forward answer. \\
S2 & Chain-of-thought & S1 + free-form reasoning before answering. \\
S3 & Self-consistency (SC@6) & S2 sampled six times with majority vote over the answers. \\
S4 & Free-text state & Explicit but unstructured world-state notes before answering. \\
S5 & Structured state & Typed physical-mental state (Section~\ref{sec:math-formalization}), no simulation. \\
S6 & \Mentis{} (full MWM) & Target observation, per-option branch simulation, value evaluation. \\
\midrule
A1 & $-$\,mental & S6 with mental state and mental observation removed. \\
A2 & $-$\,physical & S6 with physical state and physical observation removed. \\
A3 & Decoupled transition & S6 with physical and mental transitions predicted independently. \\
O1 & Oracle state & S6 with the parsed state replaced by gold $s_t^{\star}$. \\
O2 & Oracle observation & S6 with the rendered observation replaced by gold $o_t^{\target,\star}$. \\
O3 & Oracle action & S6 with action decomposition skipped; option text used verbatim. \\
O4 & Oracle transition & S6 with the simulated successor states replaced by gold ones. \\
\bottomrule
\end{tabularx}
\endgroup
\end{table}

\paragraph{Operating point.}
Reasoning effort and scoring configuration affect every downstream comparison, so they are fixed once, before any headline run, by a small calibration study on the frozen 30-record slice. All reported runs share a single global operating point, \emph{medium reasoning effort with batched comparative-rank scoring}; the calibration protocol is described in Appendix~\ref{app:operating-point}.

\paragraph{Models and roles.}
The world-model role is instantiated with eight models spanning two families~\citep{openai2023gpt4,anthropic2024claude3}, 5 OpenAI models (gpt-5.6-sol, gpt-5.5, gpt-5.4, gpt-5.4-mini, gpt-4.1) and 3 Anthropic models (claude-fable-5, claude-opus-4-8, claude-haiku-4-5), accessed through a unified endpoint. The target pseudo-agent and judge roles are held fixed, so compared systems differ only in the world model. A human reference collected under the identical task protocol provides an upper reference.

\subsection{Experimental Design}
\label{sec:exp-design}

The empirical study in Section~\ref{sec:experiments} is organized into four groups, ordered so that each builds on the ones before it.

\begin{enumerate}[label=\arabic*), leftmargin=1.6em, itemsep=1pt, topsep=2pt]
\item \textbf{Necessity ladder.} Walk the ladder of Table~\ref{tab:exp-systems} from the options-only floor (S0) through direct answering (S1) to full MWM (S6), together with the channel ablations (A1--A3), for all eight world models, and attribute outcome differences to specific modeling commitments through adjacent paired comparisons.
\item \textbf{Bottleneck localization.} Intervene on the full pipeline with gold information (O1--O4), singly and in combination: substituting the annotated state, observation, verbatim actions, or successor states into the corresponding stage measures how much error each stage contributes and which stage limits performance.
\item \textbf{Scenario study.} Slice all results by scene category and domain to measure where explicit world modeling helps most, and to check that gains depend on the type of reasoning a decision requires rather than on where it happens.
\item \textbf{Modality analysis.} Compare behavior across text, image, and sounding-video records, and intervene on the media channels (image-to-caption replacement; audio removal, frame-order shuffling, audio-only) to verify that measured performance flows through the intended evidence channels.
\end{enumerate}

Two rules apply throughout. First, the options-only floor and the channel interventions of Section~\ref{sec:metrics} are reported alongside the headline comparisons, ruling out gains that come from guessable options or leaked evidence. Second, every reported run shares the fixed operating point of Section~\ref{sec:exp-setup} and carries a complete manifest, so every number in Section~\ref{sec:experiments} can be traced to a specific configuration.

\section{Experiments and Analysis}
\label{sec:experiments}

Following the design of Section~\ref{sec:exp-design}, this section reports all the experimental results: the necessity ladder (Section~\ref{sec:exp-necessity}), oracle-based bottleneck localization (Section~\ref{sec:exp-oracle}), the scenario study (Section~\ref{sec:exp-scenario}), and the modality analysis (Section~\ref{sec:exp-media}). All numbers are final-action F1 (\%) on the full Menti-Bench data; the human reference under the same protocol is 98.5.

\subsection{Necessity Ladder}
\label{sec:exp-necessity}

\begin{figure}[!t]
\centering
\includegraphics[width=\linewidth]{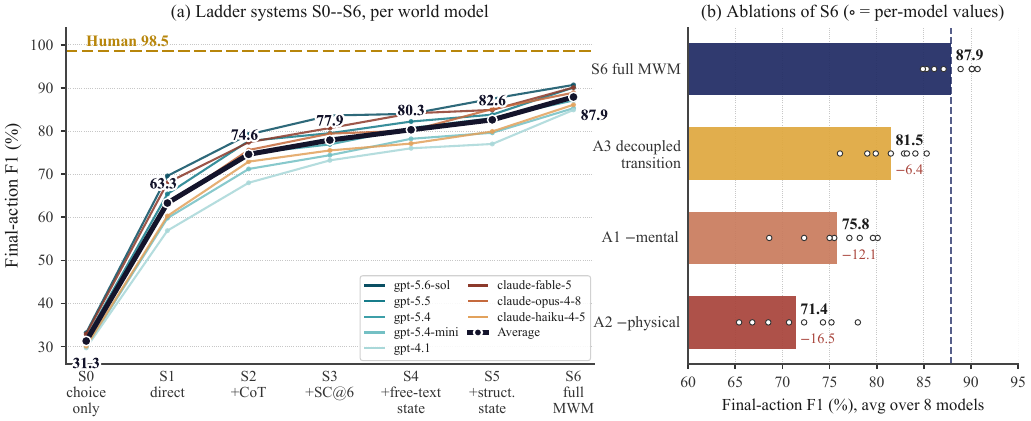}
\caption{\textbf{Necessity ladder} (final-action F1, 448 records). (a)~F1 per ladder rung for the eight world models (thin lines) and their average (thick line). (b)~Ablations of S6, averaged over the eight models; white dots mark the per-model values.}
\label{fig:exp-ladder}
\end{figure}

We run the ladder S0--S6 and the ablations A1--A3 (Table~\ref{tab:exp-systems}) with the 8 LLM-based world models. Figure~\ref{fig:exp-ladder} shows the overall pattern; Table~\ref{tab:exp-ladder} gives the full results.

\begin{table}[!t]
\caption{Necessity ladder and channel ablations: final-action F1 (\%) on all 448 records, per world model. \textbf{Avg} averages the eight models; the human reference under the identical protocol is \textbf{98.5}. Best system per column in bold.}
\label{tab:exp-ladder}
\centering
\scriptsize
\begingroup
\setlength{\tabcolsep}{4pt}
\rowcolors{4}{gray!6}{white}
\begin{tabular*}{0.98\linewidth}{@{\extracolsep{\fill}} ll r rrrrr rrr @{}}
\toprule
 & & & \multicolumn{5}{c}{\textbf{OpenAI}} & \multicolumn{3}{c}{\textbf{Anthropic}} \\
\cmidrule(lr){4-8}\cmidrule(l){9-11}
\textbf{\#} & \textbf{System} & \textbf{Avg} & \textbf{5.6-sol} & \textbf{5.5} & \textbf{5.4} & \textbf{5.4-mini} & \textbf{4.1} & \textbf{fable-5} & \textbf{opus-4-8} & \textbf{haiku-4-5} \\
\midrule
S0 & choice-only floor & 31.3 & 33.2 & 32.0 & 31.4 & 29.9 & 29.7 & 32.9 & 31.2 & 30.1 \\
S1 & direct answer & 63.3 & 69.6 & 65.4 & 63.5 & 59.8 & 56.9 & 68.0 & 62.9 & 60.3 \\
S2 & \;$+$ CoT & 74.6 & 79.2 & 77.7 & 74.8 & 71.2 & 68.0 & 77.4 & 75.6 & 72.9 \\
S3 & \;$+$ SC@6 & 77.9 & 83.6 & 79.5 & 76.9 & 74.4 & 73.2 & 80.7 & 79.4 & 75.5 \\
S4 & \;$+$ free-text state & 80.3 & 84.0 & 82.2 & 80.7 & 78.2 & 76.0 & 84.1 & 80.1 & 77.1 \\
S5 & \;$+$ structured state & 82.6 & 87.5 & 83.8 & 83.0 & 79.6 & 77.0 & 84.9 & 85.1 & 79.9 \\
S6 & \textbf{full MWM (\Mentis{})} & \textbf{87.9} & \textbf{90.7} & \textbf{90.1} & \textbf{87.1} & \textbf{85.3} & \textbf{84.9} & \textbf{90.1} & \textbf{88.9} & \textbf{86.1} \\
\midrule
A1 & $-$\,mental & 75.8 & 80.1 & 78.2 & 75.5 & 72.3 & 68.6 & 79.6 & 77.1 & 75.0 \\
A2 & $-$\,physical & 71.4 & 78.0 & 74.3 & 70.7 & 66.8 & 65.4 & 75.2 & 72.3 & 68.5 \\
A3 & decoupled transition & 81.5 & 85.3 & 84.1 & 82.9 & 79.0 & 76.1 & 83.2 & 81.5 & 79.9 \\
\bottomrule
\end{tabular*}
\endgroup
\end{table}

\paragraph{F1 increases at every rung, for every model.}
The average rises monotonically from 31.3 (S0) to 87.9 (S6), and the ordering holds for each of the models. The two largest increments come from reading the story ($+32.0$, S0$\rightarrow$S1) and from explicit reasoning ($+11.3$, S1$\rightarrow$S2); the modeling rungs then add $+2.4$, $+2.3$, and $+5.3$. This proves our central claim that \coreclaim{every added commitment of mental world modeling improves target-action prediction, and the full MWM pipeline is the best configuration for all world models.}

\paragraph{The options-only floor is low and flat.}
S0 lies within 29.7--33.2 for all models, about 32 points below S1. The band is above random guessing ($1/6\approx16.7$) because some distractors are semantically less plausible next actions, but it is nearly constant across model capability: a stronger model extracts nothing more from the options alone. The conclusion is that Menti-Bench cannot be solved from the option set; the scores in this section measure story understanding, as intended.

\begin{figure}[!t]
\centering
\includegraphics[width=\linewidth]{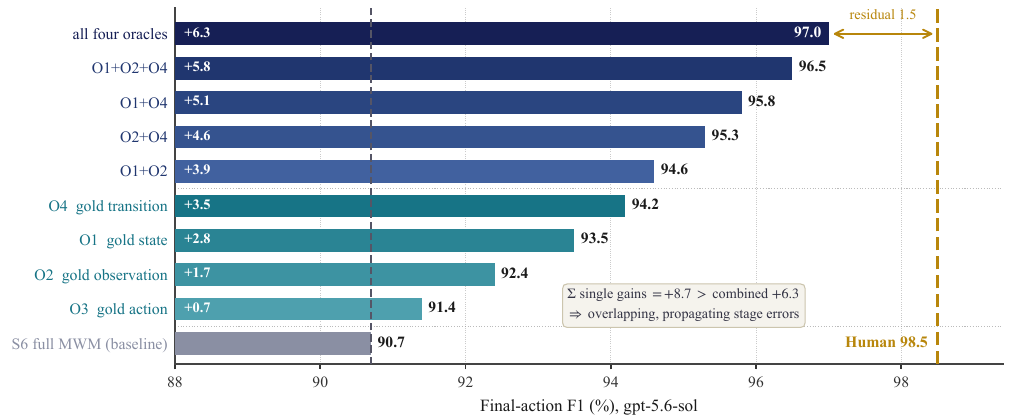}
\caption{\textbf{Oracle interventions} (gpt-5.6-sol). Bars show final-action F1 when gold information replaces predicted artifacts at one or more stages; bar labels give the gain over fully predictive S6.}
\label{fig:exp-oracle}
\end{figure}

\paragraph{Direct answering does not suffice, even with more compute.}
S3 samples six reasoning chains per record and still trails S6 by 10.0 on average, and S6 with the weakest model (gpt-4.1, 84.9) exceeds S3 with the strongest model (gpt-5.6-sol, 83.6). The gap between direct answering and full MWM is therefore not an artifact of computation budget: \coreclaim{world-model-style simulation provides gains that direct answering cannot reach by sampling more.}

\paragraph{Explicit state representation helps before any simulation.}
Writing an unformatted world-state note (S4, 80.3) already exceeds six-sample self-consistency (S3, 77.9), and typing the state into the physical-mental schema (S5, 82.6) adds further. The largest modeling increment ($+5.3$) then comes from the final rung, observation rendering, branch simulation, and value evaluation. The conclusion is that both halves of \Mentis{} contribute: the state representation itself, and the simulation machinery built on top of it.

\paragraph{Both channels and their coupling are necessary.}
Removing the mental channel (A1) costs 12.1 points on average, removing the physical channel (A2) costs 16.5, and predicting the two transitions independently (A3) costs 6.4; the ordering S6 $>$ A3 $>$ A1 $>$ A2 holds for all eight models. This verifies the three structural claims of the framework: \coreclaim{physical modeling alone is insufficient (mental state is necessary), mental reasoning degrades without physical grounding, and the joint physical-mental transition outperforms modeling the two channels separately.}

\paragraph{The gain over direct answering is larger for weaker base models.}
Stronger base models score higher at every rung, but the S6$-$S1 gain moves in the opposite direction: $+21.1$ on gpt-5.6-sol vs.\ $+28.0$ on gpt-4.1, and $+22.1$ on claude-fable-5 vs.\ $+25.8$ on claude-haiku-4-5. The two families show the same rung and ablation orderings, so the pattern is not specific to one family. Explicit structure is therefore not made redundant by stronger base models; it benefits all tiers and weaker ones most. Finally, humans reach 98.5 under the identical protocol, so the task itself is nearly unambiguous; the remaining 7.8-point gap of the best configuration is a modeling shortfall, and the next subsection localizes it.

\subsection{Bottleneck Localization}
\label{sec:exp-oracle}

An oracle intervention replaces one predicted intermediate artifact with its gold annotation (O1 state, O2 observation, O3 action, O4 transition; Table~\ref{tab:exp-systems}) while leaving the other stages predictive. The gain of a single oracle measures the error contributed by that stage; combinations measure how much of the remaining gap gold upstream information can close. All runs use gpt-5.6-sol (Figure~\ref{fig:exp-oracle}, Table~\ref{tab:exp-oracle}).

\begin{table}[!t]
\caption{Oracle interventions on S6 (gpt-5.6-sol): single stages and combinations. $\Delta$ is the gain over fully predictive S6 (90.7); the human reference is 98.5.}
\label{tab:exp-oracle}
\centering
\small
\begingroup
\rowcolors{2}{gray!6}{white}
\begin{tabular*}{0.9\linewidth}{@{\extracolsep{\fill}} llrrr @{}}
\toprule
\textbf{\#} & \textbf{Configuration} & \textbf{F1} & \textbf{$\Delta$ vs.\ S6} & \textbf{Gap to human} \\
\midrule
S6 & full MWM, all stages predicted & 90.7 & -- & 7.8 \\
\midrule
O1 & gold state & 93.5 & $+2.8$ & 5.0 \\
O2 & gold observation & 92.4 & $+1.7$ & 6.1 \\
O3 & gold action (skip decomposition) & 91.4 & $+0.7$ & 7.1 \\
O4 & gold transition & \textbf{94.2} & $\bm{+3.5}$ & 4.3 \\
\midrule
O1$+$O2 & state $+$ observation & 94.6 & $+3.9$ & 3.9 \\
O2$+$O4 & observation $+$ transition & 95.3 & $+4.6$ & 3.2 \\
O1$+$O4 & state $+$ transition & 95.8 & $+5.1$ & 2.7 \\
O1$+$O2$+$O4 & state $+$ observation $+$ transition & 96.5 & $+5.8$ & 2.0 \\
O1--O4 & all four stages & \textbf{97.0} & $\bm{+6.3}$ & \textbf{1.5} \\
\bottomrule
\end{tabular*}
\endgroup
\end{table}

\begin{figure}[!t]
\centering
\includegraphics[width=\linewidth]{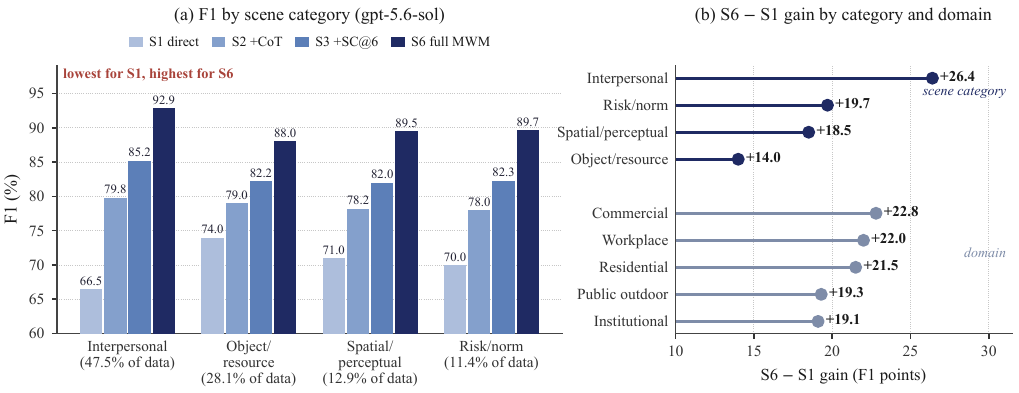}
\caption{\textbf{Scenario study} (gpt-5.6-sol). (a)~F1 by scene category for four systems. (b)~S6$-$S1 gain by scene category and by domain.}
\label{fig:exp-scenario}
\end{figure}

\begin{table}[!t]
\caption{Final-action F1 (\%) by scene category and domain (gpt-5.6-sol, 448 records). Column shares give each slice's fraction of the data; the last row is the S6$-$S1 gain.}
\label{tab:exp-scenario}
\centering
\scriptsize
\begingroup
\setlength{\tabcolsep}{3.5pt}
\rowcolors{4}{gray!6}{white}
\begin{tabular*}{0.98\linewidth}{@{\extracolsep{\fill}} l r rrrr rrrrr @{}}
\toprule
 & & \multicolumn{4}{c}{\textbf{Scene category}} & \multicolumn{5}{c}{\textbf{Domain}} \\
\cmidrule(lr){3-6}\cmidrule(l){7-11}
\textbf{System} & \textbf{Overall} & \textbf{Interp.} & \textbf{Obj./res.} & \textbf{Spatial} & \textbf{Risk/norm} & \textbf{Resid.} & \textbf{Outdoor} & \textbf{Commer.} & \textbf{Institut.} & \textbf{Workpl.} \\
 & & {\tiny 47.5\%} & {\tiny 28.1\%} & {\tiny 12.9\%} & {\tiny 11.4\%} & {\tiny 52.0\%} & {\tiny 14.7\%} & {\tiny 12.1\%} & {\tiny 11.6\%} & {\tiny 9.6\%} \\
\midrule
S1 direct & 69.6 & 66.5 & 74.0 & 71.0 & 70.0 & 69.3 & 70.6 & 68.7 & 71.0 & 69.0 \\
S2 $+$CoT & 79.2 & 79.8 & 79.0 & 78.2 & 78.0 & 79.5 & 78.4 & 80.1 & 79.0 & 77.8 \\
S3 $+$SC@6 & 83.6 & 85.2 & 82.2 & 82.0 & 82.3 & 83.4 & 84.8 & 82.9 & 84.1 & 83.2 \\
S6 full MWM & \textbf{90.7} & \textbf{92.9} & \textbf{88.0} & \textbf{89.5} & \textbf{89.7} & \textbf{90.8} & \textbf{89.9} & \textbf{91.5} & \textbf{90.1} & \textbf{91.0} \\
\midrule
\rowcolor{white}
S6$-$S1 gain & $+21.1$ & $+26.4$ & $+14.0$ & $+18.5$ & $+19.7$ & $+21.5$ & $+19.3$ & $+22.8$ & $+19.1$ & $+22.0$ \\
\bottomrule
\end{tabular*}
\endgroup
\end{table}

\paragraph{Transition simulation is the largest single bottleneck.}
Gold transitions gain $+3.5$, gold states $+2.8$, gold observations $+1.7$, and skipping action decomposition $+0.7$. The transition oracle alone recovers 45\% of the 7.8 human gap. Action decomposition contributes little remaining error; the option text is already usable as given.

\paragraph{Oracle gains are sub-additive.}
The four single gains sum to $+8.7$, but all four oracles together gain $+6.3$. Stage errors overlap: an upstream error, such as a mis-parsed state, also produces a downstream one, such as a wrong simulated consequence, and is then counted by both oracles. Every combination containing O4 outperforms those without it, consistent with the single-oracle ranking. The combined gain, not the sum of single gains, is the correct estimate of the recoverable error.

\paragraph{Most of the remaining gap comes from the intermediate stages.}
With all four artifacts gold, the pipeline reaches 97.0. Of the 7.8-point human gap, 6.3 points (81\%) are attributable to prediction errors in the intermediate stages, mainly transition simulation and state parsing, and 1.5 points (19\%) remain in value evaluation, the decision rule, or residual item difficulty. The conclusion, and the direction for future work: \coreclaim{what limits current MWM is simulating how the coupled world changes, not representing what it currently is; improvements should target the transition model first.}

\subsection{Scenario Study}
\label{sec:exp-scenario}

The framework predicts that gains should depend on the type of reasoning a decision requires rather than on where the scene takes place. We slice all 448 records by scene category and by domain, comparing S1, S2, S3, and S6 on gpt-5.6-sol (Figure~\ref{fig:exp-scenario}, Table~\ref{tab:exp-scenario}).

\paragraph{Gains are largest on interpersonal scenes.}
Interpersonal decisions are the weakest category for direct answering (66.5, below S1's overall 69.6) and the strongest for full MWM (92.9), giving the largest gain ($+26.4$). Object/resource scenes show the smallest gain ($+14.0$): their higher S1 score (74.0) indicates that they rely more on commonsense affordances that direct answering already captures. The between-category spread narrows from 7.5 points under S1 to 4.9 under S6, with all four categories at or above 88.0. This is the necessity claim confirmed at the slice level: \coreclaim{the benefit of mental world modeling is largest exactly where hidden mental variables govern the decision.}

\paragraph{Gains are flat across domains.}
The five domain gains lie within 19.1--22.8. This matches the prediction: domains change the setting but not the type of reasoning required, so the gains are modeling gains rather than surface-pattern gains. A spike in one domain would instead have suggested surface regularities in the data.

\subsection{Modality Analysis}
\label{sec:exp-media}

\begin{figure}[!t]
\centering
\includegraphics[width=.98\linewidth]{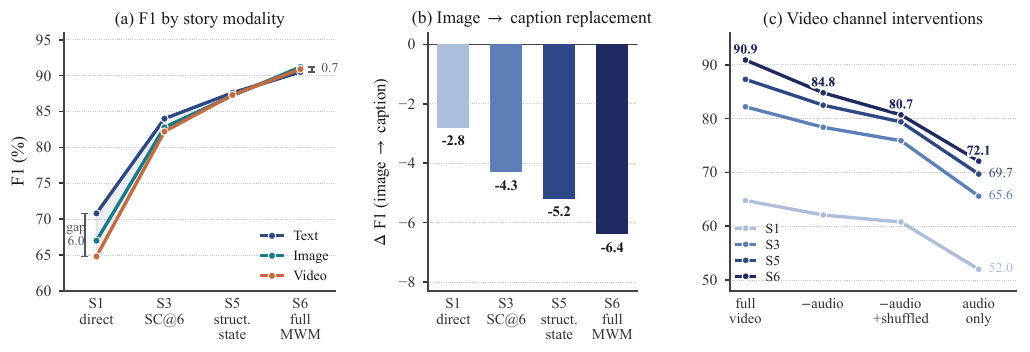}
\vspace{-2mm}
\caption{\textbf{Modality analysis} (gpt-5.6-sol). (a)~F1 by story modality per ladder system. (b)~Effect of replacing images with captions. (c)~F1 under video channel interventions.}
\label{fig:exp-modality}
\end{figure}

We first slice four ladder systems by story modality, then intervene on the media channels: replacing images with neutral captions, removing video audio, additionally shuffling frame order, and keeping only audio. All runs use gpt-5.6-sol (Figure~\ref{fig:exp-modality}; Tables~\ref{tab:exp-modality} and~\ref{tab:exp-channels}).

\begin{table}[!t]
\caption{Final-action F1 (\%) by story modality (gpt-5.6-sol).}
\label{tab:exp-modality}
\centering
\small
\begingroup
\rowcolors{2}{gray!6}{white}
\begin{tabular*}{0.82\linewidth}{@{\extracolsep{\fill}} lrrrr @{}}
\toprule
\textbf{System} & \textbf{All (448)} & \textbf{Text (320)} & \textbf{Image (100)} & \textbf{Video (28)} \\
\midrule
S1 direct & 69.6 & 70.8 & 67.0 & 64.8 \\
S3 SC@6 & 83.6 & 84.0 & 82.8 & 82.2 \\
S5 structured state & 87.5 & 87.6 & 87.2 & 87.3 \\
S6 full MWM & \textbf{90.7} & \textbf{90.5} & \textbf{91.2} & \textbf{90.9} \\
\midrule
\rowcolor{white}
S6$-$S1 gain & $+21.1$ & $+19.7$ & $+24.2$ & $+26.1$ \\
\bottomrule
\end{tabular*}
\endgroup
\end{table}

\begin{table}[!t]
\caption{Channel interventions (gpt-5.6-sol): image records with images replaced by neutral captions; video records with audio removed, frame order additionally shuffled, or only audio retained.}
\label{tab:exp-channels}
\vspace{-2mm}
\centering
\small
\begingroup
\setlength{\tabcolsep}{4pt}
\rowcolors{3}{gray!6}{white}
\begin{tabular*}{0.98\linewidth}{@{\extracolsep{\fill}} l rrr rrrr @{}}
\toprule
 & \multicolumn{3}{c}{\textbf{Image (n$=$100)}} & \multicolumn{4}{c}{\textbf{Video (n$=$28)}} \\
\cmidrule(lr){2-4}\cmidrule(l){5-8}
\textbf{System} & \textbf{Original} & \textbf{$\to$ caption} & \textbf{$\Delta$} & \textbf{Full} & \textbf{$-$audio} & \textbf{$-$audio$+$shuf.} & \textbf{Audio only} \\
\midrule
S1 direct & 67.0 & 64.2 & $-2.8$ & 64.8 & 62.1 & 60.8 & 52.0 \\
S3 SC@6 & 82.8 & 78.5 & $-4.3$ & 82.2 & 78.4 & 75.9 & 65.6 \\
S5 structured state & 87.2 & 82.0 & $-5.2$ & 87.3 & 82.5 & 79.4 & 69.7 \\
S6 full MWM & 91.2 & 84.8 & $\bm{-6.4}$ & 90.9 & 84.8 & 80.7 & 72.1 \\
\bottomrule
\end{tabular*}
\endgroup
\vspace{-3mm}
\end{table}

\paragraph{Structure closes the modality gap.}
Direct answering is lower on media inputs: 70.8 on text, 67.0 on image, 64.8 on video. The text-video gap shrinks ($6.0 \rightarrow 1.8 \rightarrow 0.3$) and is gone under S6 (90.5, 91.2, 90.9); correspondingly, the S6$-$S1 gain is largest on video ($+26.1$) and image ($+24.2$). The gap already closes at S5, before any simulation: once the story is parsed into the typed state, the downstream stages no longer depend on the input modality. \coreclaim{The advantage of mental world modeling is modality-general: the structured state converts text, visual, and audio evidence into a common format, and the media penalty of direct answering disappears.}

\paragraph{Channel interventions confirm that the media evidence is used.}
Every intervention lowers F1, and the loss grows with the amount of structure in all sixteen system-intervention cells: replacing images with captions costs S1 2.8 points and S6 6.4; removing audio costs S1 2.7 and S6 6.1; shuffling the remaining frames costs S6 a further 4.1; keeping only audio costs S6 18.8. A system answering mainly from textual priors would be largely unaffected by these interventions, so the losses indicate that the media channels carry real evidence and that the structured pipeline uses more of it than direct answering does. For S6, the visual stream matters most ($-18.8$), then audio ($-6.1$), then frame order ($-4.1$), consistent with the sounding-video design in which audio carries part but not all of the evidence. Given the subset sizes (100 image, 28 video), we treat the absolute deltas as directional; the monotone pattern across systems is the stable result. The conclusion is that the modality-general gains above rest on genuine use of the visual, temporal, and auditory evidence, not on textual priors.

\section{Conclusion}
\label{sec:conclusion}

This paper introduces Mental World Modeling as a framework for world models that reason about people rather than only objects. The central move is to treat physical and mental states as one coupled world state, render a target-specific partial observation from that state, and simulate how a candidate action changes both the material scene and the mental-social configuration. We formalize this idea, specify a practical state taxonomy, implement \Mentis{} as an inspectable baseline, and evaluate the necessity and limits of MWM on a process-annotated testbed. The evidence is consistent across 8 modern LLM-based world models from two families: explicit mental world modeling is necessary for predicting human decisions, the mental channel, the physical channel, and their coupled transition are all required, and the gains are largest on interpersonal scenes, where hidden mental variables govern the decision, while holding across text, image, and sounding-video inputs. Oracle interventions localize most of the remaining gap to the human reference in transition simulation, which gives future improvement a clear priority order. The larger claim is that human-centered AI requires world models whose latent variables include beliefs, goals, intentions, emotions, norms, relationships, and social atmosphere. Without these variables, many apparently simple human actions remain unpredictable for the wrong reason: the model looks at the world, but not at the world as the target agent understands it.

{
  \small
  \bibliographystyle{unsrtnat}
  \bibliography{ref}
}

\newpage
\appendix
\addtocontents{toc}{\protect\setcounter{tocdepth}{-10}}%
\appendixtoc
\startappendixtoccapture

\newpage
\section{Ethics and Responsible Use}
\label{app:ethics}

MWM makes assumptions about what people believe, want, and feel explicit enough to be simulated and tested. That is the source of its scientific value and of its risks, so we state the ethical position of this work in full.

\paragraph{What the framework does and does not claim.}
MWM is an external, approximate, task-relevant simulator of mental and social variables. It does not simulate private conscious experience, and none of our results should be read as evidence that a language model \emph{has} a Theory of Mind or understands people in a human sense. The claim tested in this paper is functional: making mental variables explicit improves the prediction of target actions and makes the prediction process auditable. Anthropomorphic readings of the scores are not supported by the experiments and should be avoided.

\paragraph{Mental-state inference is hypothesis, not measurement.}
Every mental variable a system fills in is an inference from behavior and context, not an observation. Systems built on this framework should represent such variables with uncertainty, restrict them to the scope the task requires, prefer asking or deferring when confidence is low, and keep them contestable: a person should be able to see, question, and correct what a system assumed about them. The framework supports this discipline, since every assumed state is explicit and auditable, but it does not enforce it; builders remain responsible.

\paragraph{Privacy and data protection.}
The benchmark contains no personal data: all characters, scenes, and dialogues are fictional, and all images and videos are synthetically generated. No real person's behavior, likeness, or communications were collected. Applied to real users, however, the same machinery would constitute processing of sensitive information (beliefs, emotions, vulnerabilities) and would require informed consent, purpose limitation, data minimization, retention limits, and compliance with applicable data-protection law. Mental-state inference must not be used for covert profiling or monitoring.

\paragraph{Manipulation and dual use.}
An explicit model of what a person believes, fears, and can be persuaded of supports assistance and manipulation alike; intent and governance separate the two uses. We consider the following out of scope and inappropriate: targeted persuasion or deception built on inferred vulnerabilities, emotional manipulation, covert influence, surveillance of employees or students, and any use that treats inferred mental states as grounds for punitive decisions. The design choice that most directly mitigates misuse is legibility: \Mentis{} exposes every assumed state, observation, and simulated consequence as an inspectable artifact, so a misapplied inference leaves a trail that can be audited and contested.

\paragraph{High-stakes and vulnerable contexts.}
Care, health, education, and legal settings involve people who may be least able to contest an incorrect mental-state inference. In such settings, MWM-style components should operate under human oversight, support escalation and deferral rather than autonomous judgment, and must not be used for psychological assessment or diagnosis; nothing in this paper validates them for that purpose.

\paragraph{Bias and cultural scope.}
Norms, politeness conventions, emotional expression, and role expectations are culturally situated. The benchmark is authored in English around everyday residential, public, commercial, institutional, and workplace scenes, and its norm judgments reflect the conventions its authors and annotators considered defensible; they are not universal. Systems evaluated or tuned on it inherit this scope, and applying them across cultural contexts requires recalibrating the normative layer, not just translating the text.

\paragraph{Human involvement and synthetic media.}
Benchmark construction involved human authoring, auditing, and adjudication of fictional everyday scenes; the material contains no graphic or traumatic content, and no crowdsourced psychological data was collected. The human reference of Section~\ref{sec:benchmark} was collected under the identical task protocol on the same fictional records. All media assets are generated and quality-controlled; they depict fictional characters and are not intended to resemble real individuals.

\paragraph{Release and contamination.}
The benchmark is released for research evaluation only. Its gold process annotations are the evaluation target, so they must be kept out of training corpora: training on them defeats the diagnostic purpose of the data and contaminates exactly the process signals the suite measures. We ask users not to redistribute the gold blocks in plain text through training-scrapable channels.

\section{Extended Description of \Mentis}
\label{app:mentis-extended}

\subsection{Inspectability, Ablations, and Experimental Use}
\label{sec:mentis-inspectability}

\Mentis{} is designed as an experimental scaffold. Each run creates a timestamped directory containing predictions, raw prompts, raw model responses, parsed JSON artifacts, a run manifest, LLM traces, performance summaries, and a lightweight performance diagram. This logging is not merely engineering hygiene. It makes MWM testable at the level where the theory makes claims: state representation, target observation, action semantics, physical transition, mental transition, and value evaluation.

\begin{table}[H]
\caption{Ablation views supported by the current \Mentis{} implementation. These variants are intended for mechanism analysis rather than as independent systems.}
\label{tab:mentis-ablations}
\centering
\footnotesize
\setlength{\tabcolsep}{4pt}
\renewcommand{\arraystretch}{1.1}
\rowcolors{2}{gray!7}{white}
\begin{tabularx}{\linewidth}{
>{\raggedright\arraybackslash}p{0.32\linewidth}
>{\raggedright\arraybackslash}p{0.31\linewidth}
>{\raggedright\arraybackslash}X}
\toprule
\rowcolor{white}
\textbf{Variant} & \textbf{Intervention} & \textbf{Question tested} \\
\midrule
\texttt{full\_mwm} & Uses full state, target observation, coupled transitions, and branch scoring. & How well does the complete baseline instantiate MWM? \\
\texttt{no\_mental\_information} & Hides mental state and mental observation from evaluation. & How much does mental information contribute to final choice? \\
\texttt{no\_physical\_information} & Hides physical state and physical observation from evaluation. & Which failures arise when physical grounding is removed? \\
\texttt{decoupled\_transition} & Predicts physical and mental transitions independently, without coupling. & Does joint physical-mental evolution outperform separate prediction? \\
\texttt{oracle\_state} & Replaces parsed state with annotated gold state. & How much error comes from state parsing? \\
\texttt{oracle\_observation} & Replaces generated target observation with annotated gold observation. & How much error comes from perspective rendering? \\
\texttt{oracle\_action} & Skips action decomposition and uses option text verbatim. & Does action decomposition contribute error? \\
\texttt{oracle\_transition} & Replaces simulated successor states with annotated gold ones. & How much error comes from consequence simulation? \\
\texttt{direct\_answer\_baseline} & Bypasses the MWM chain and asks for an option directly (with CoT, self-consistency, free-text-state, and structured-state variants). & Does explicit world modeling outperform direct answering? \\
\bottomrule
\end{tabularx}
\rowcolors{2}{}{}
\renewcommand{\arraystretch}{1.0}
\end{table}

The most important use of these variants (Table~\ref{tab:mentis-ablations}) is not to compute aggregate accuracy alone. A full MWM run may fail because the state parser invents a belief, because the observation generator leaks hidden information, because the mental transition ignores a social obligation, or because the evaluator overweights physical feasibility. The oracle and information-removal settings separate these explanations, supporting the experimental agenda of Section~\ref{sec:experiments}.

\paragraph{Scoring and decision details.}
Batched comparative scoring is the default: all successful branches are graded within the same prompt context, and the evaluator returns per-dimension grades plus a strict comparative ranking. If batched scoring fails schema validation, \Mentis{} falls back to individual branch scoring. The decision module selects the highest-value branch and resolves ties by a fixed cascade, comparative rank first, then the per-dimension scores, then a seeded deterministic fallback, kept intentionally outside the language model. The raw model outputs, parsed score objects, normalized values, and the decision trace are persisted with every run.

\subsection{Multimodal Front-End}
\label{app:mentis-multimodal}

The state parser is modality-aware, building on instruction-following multimodal LLMs~\citep{liu2023visual}, and the media ingestion path is part of the measured system rather than hidden preprocessing.

\paragraph{Image stories.}
Image records supply one to five images together with a short scene-anchor text. The parser prompt instructs the model to treat caption boxes inside images as narration to be quoted verbatim, to establish temporal order across the image sequence, and to track visibility asymmetries (who can see what), since several records hinge on them.

\paragraph{Video stories: adaptive frame sampling.}
Because the underlying models ingest frames rather than continuous video, the front-end samples frames adaptively: scene-change detection proposes candidate timestamps, a uniform anchor grid guarantees coverage of quiet segments, near-duplicate timestamps are merged under a minimum gap, and the sequence is capped at a frame budget. Each frame is passed with its timestamp so that the parser can reason about event order. A uniform-sampling fallback and the sampling parameters are exposed in configuration, which is what enables the front-end ablation of Section~\ref{sec:exp-media}.

\paragraph{Video stories: audio channel.}
The benchmark's videos are \emph{sounding} videos: dialogue and ambient audio carry evidence that frames alone miss. The front-end extracts the audio track, transcribes it with a speech-recognition model~\citep{radford2023robust}, and injects the transcript alongside the frames, with per-file caching and graceful degradation (a transcription failure annotates the run metadata instead of aborting the sample). The parser prompt attributes transcript lines to speakers where the frames make the speaker identifiable.

\subsection{Limitations of the Baseline}
\label{sec:mentis-limitations}

\Mentis{} should be read as a baseline implementation, not as the final architecture of Mental World Modeling. It inherits the weaknesses of prompted LLM/MLLM modules: sensitivity to prompt wording, schema failures, over-inference of mental states, and error propagation across stages. Its current benchmark mode also constrains the action set to annotated options, so it does not yet evaluate open-ended policy generation. The one-step transition format is appropriate for early component analysis, but longer-horizon MWM will require memory, uncertainty propagation, model calibration, and learned transition functions.

\section{Theoretical Grounding of MWM}
\label{app:theory-grounding}

Table~\ref{tab:mwm-theory-grounding} collects the arguments of Section~\ref{sec:theory-grounding}, pairing each theoretical source with the MWM commitment it supports.

\begin{table}[H]
\caption{Theoretical grounding for MWM. Each theory supports a specific modeling commitment used in the formal framework.}
\label{tab:mwm-theory-grounding}
\centering
\small
\rowcolors{2}{gray!7}{white}
\begin{tabularx}{\linewidth}{p{0.24\linewidth}p{0.32\linewidth}X}
\toprule
\rowcolor{white}
\textbf{Theoretical source} & \textbf{Core idea} & \textbf{Implication for MWM} \\
\midrule
Model-based cognition and human-like learning~\citep{craik1943nature,lake2017building}
& Intelligence uses structured models to evaluate possible futures; human-like learning requires intuitive physics and psychology.
& MWM must model both physical evolution and agent-level mental dynamics. \\
Theory of Mind and BDI~\citep{premack1978chimpanzee,bratman1987intention,rao1995bdi}
& Beliefs, desires, and intentions are latent variables that explain and predict behavior.
& Mental variables are causal state variables, not post-hoc textual rationales. \\
Active inference~\citep{friston2010free}
& Internal and external states are coupled through perception and action.
& MWM needs partial observation and joint physical-mental state transition. \\
Embodied/enactive cognition~\citep{varela1991embodied}
& Cognition is generated through situated interaction with the environment.
& Mental state must be interpreted inside physical and social context. \\
Ecological affordances~\citep{gibson1979ecological}
& Agents perceive action possibilities, not only geometry.
& Observation generation should render target-specific affordances and social meanings. \\
POMDPs~\citep{kaelbling1998planning}
& Agents act under partial observability of a hidden state.
& The target's first-person observation differs from the MWM's third-person joint state. \\
\bottomrule
\end{tabularx}
\rowcolors{2}{}{}
\end{table}

\section{Full State Taxonomy}
\label{app:state-taxonomy}

Table~\ref{tab:state-taxonomy-full-physical} and Table~\ref{tab:state-taxonomy-full-mental} expand the compact schema in Section~\ref{sec:state-taxonomy}. The entries mirror the annotation guidelines used during benchmark construction.

\begin{table}[H]
\caption{Full physical-state taxonomy for MWM. Repeated Level-1 and Level-2 cells are merged, and rows alternate shading to aid row-by-row reading.}
\label{tab:state-taxonomy-full-physical}
\centering
\footnotesize
\setlength{\tabcolsep}{4pt}
\renewcommand{\arraystretch}{1.2}
\rowcolors{2}{gray!9}{white}
\begin{tabularx}{\linewidth}{@{}>{\raggedright\arraybackslash}p{0.15\linewidth}>{\raggedright\arraybackslash}p{0.17\linewidth}>{\raggedright\arraybackslash}p{0.185\linewidth}>{\raggedright\arraybackslash}X@{}}
\toprule
\rowcolor{white}
\textbf{Level 1} & \textbf{Level 2} & \textbf{Level 3} & \textbf{Example fields} \\
\midrule
 &  & Intrinsic properties & name, size, weight, color, material, appearance \\
 & \multirow{-2}{*}{Objects} & Contextual attributes & motion state, position, physical condition, semantic content \\
\cmidrule(l){2-4}
 &  & Intrinsic properties & height, weight, clothing type and color, appearance \\
\multirow{-4}{*}{\makecell[l]{Entity and\\ attribute}} & \multirow{-2}{*}{Characters} & Contextual attributes & facial expression, pose, gesture, gaze, motion state, position, body condition, speech \\
\midrule
 &  & Relative position & object-object, person-person, object-person relative location \\
 & \multirow{-2}{*}{\makecell[l]{Spatial\\ relations}} & Distance and occlusion & closeness, partial blocking, visibility constraints \\
\cmidrule(l){2-4}
\multirow{-3}{*}{Relations} & Contact relations & Physical contact & person-object and person-person contact \\
\midrule
 & Location & Time, place, region & classroom, home, street, hospital, office, day/night \\
\multirow{-2}{*}{Environment} & \makecell[l]{Ambient\\ condition} & Visual, acoustic, air, space & lighting, visibility, noise, background sound, temperature, wind, humidity, smoke, crowdedness, clutter, openness \\
\bottomrule
\end{tabularx}
\end{table}

\begin{table}[H]
\caption{Full mental-state taxonomy for MWM. Repeated Level-1 and Level-2 cells are merged, and rows alternate shading to aid row-by-row reading.}
\label{tab:state-taxonomy-full-mental}
\centering
\footnotesize
\setlength{\tabcolsep}{4pt}
\renewcommand{\arraystretch}{1.2}
\rowcolors{2}{gray!9}{white}
\begin{tabularx}{\linewidth}{@{}>{\raggedright\arraybackslash}p{0.15\linewidth}>{\raggedright\arraybackslash}p{0.17\linewidth}>{\raggedright\arraybackslash}p{0.2\linewidth}>{\raggedright\arraybackslash}X@{}}
\toprule
\rowcolor{white}
\textbf{Level 1} & \textbf{Level 2} & \textbf{Level 3} & \textbf{Example fields} \\
\midrule
 &  & Identity attributes & name, occupation, group type, role \\
 &  & Epistemic state & beliefs, attention focus, knowledge, ignorance, false belief \\
 &  & Motivational state & goals, intentions, plans, desires \\
 &  & Affective state & emotions, mood, anxiety, anger, relief \\
 &  & Dispositional state & preferences, values, personality, habits \\
\multirow{-6}{*}{\makecell[l]{Mental entity\\ and attribute}} & \multirow{-6}{*}{\makecell[l]{Individual\\ / group}} & Normative state & rules, cultural norms, customs, obligations, prohibitions \\
\midrule
 & Attitudes & Person-object, person-person, person-group, group-group & likes, distrusts, respects, fears, owes, avoids \\
\multirow{-2}{*}{Relations} & Role relations & Person-person, person-group, group-group & teacher-student, doctor-patient, parent-child, customer-staff \\
\midrule
Environment & Atmosphere & Social atmosphere & tense, cooperative, hostile, festive, awkward, urgent \\
\bottomrule
\end{tabularx}
\end{table}

\section{Module Interface Schemas}
\label{app:schemas}

The modules of \Mentis{} communicate through typed JSON objects. Making these schemas explicit is what allows every intermediate artifact to be validated, diffed against gold, and ablated. The listings below give the abridged contracts; optional fields are omitted for readability, and the full field set follows the taxonomies of Appendix~\ref{app:state-taxonomy}.

\paragraph{Joint world state $s_t$ (\textsc{StateParser} output).}
\begin{lstlisting}[style=json,basicstyle=\ttfamily\scriptsize]
{
  "physical_state": {
    "objects":    [{"id": "", "name": "", "attributes": {}, "position": "",
                    "motion_state": "", "semantic_content": ""}],
    "characters": [{"id": "", "name": "", "appearance": "", "pose": "", "gaze": "",
                    "position": "", "motion_state": "", "speech": ""}],
    "relations":  {"spatial": [], "contact": []},
    "environment":{"time": "", "place": "", "region": "",
                   "ambient": {"visual": "", "acoustic": "", "air": "", "space": ""}}
  },
  "mental_state": {
    "individuals": [{"id": "", "identity": "", "beliefs": "", "attention": "",
                     "goals": "", "intentions": "", "emotions": "",
                     "dispositions": "", "norms": "", "constraints": ""}],
    "groups":      [{"id": "", "identity": "", "beliefs": "", "goals": "",
                     "emotions": "", "norms": "", "constraints": ""}],
    "relations":   {"attitudes": [], "role_relations": []},
    "atmosphere":  ""
  }
}
\end{lstlisting}

\paragraph{Target observation $o_t^{\target}$ (\textsc{ObservationGenerator} output).}
Fields the target cannot access are set to the literal \texttt{"Unknown"}; higher-order entries carry the inference order $\ell$.
\begin{lstlisting}[style=json,basicstyle=\ttfamily\scriptsize]
{
  "target_agent": "",
  "physical_observation": {"visible": [], "audible": [], "self_state": ""},
  "mental_observation": {
    "self":  {"beliefs": "", "goals": "", "emotions": ""},
    "others":[{"agent": "", "order": 1, "inferred_beliefs": "",
               "inferred_intentions": "", "inferred_emotions": ""}]
  }
}
\end{lstlisting}

\paragraph{Candidate action $a_{t,k}^{\target}$ (\textsc{ActionParser} output).}
\begin{lstlisting}[style=json,basicstyle=\ttfamily\scriptsize]
{
  "option_id": "A",
  "physical_action": "",
  "mental_action": "",
  "affected_objects": []
}
\end{lstlisting}

\paragraph{Branch score $r_k$ (\textsc{ScoringModule} output).}
\begin{lstlisting}[style=json,basicstyle=\ttfamily\scriptsize]
{
  "option_id": "A",
  "physically_plausible": 0.0,
  "mentally_consistent": 0.0,
  "socially_appropriate": 0.0,
  "relative_rank": 1,
  "veto": false,
  "reasoning": ""
}
\end{lstlisting}

\section{Menti-Bench: Construction, Composition, and Statistics}
\label{app:benchmark-details}

This appendix expands the summary description of Section~\ref{sec:dataset}: the conversion pipeline and per-modality quality control, the composition and summary statistics of the released records, and the evaluation objects the testbed exposes.
The small size of the data supports the uses made of it in this paper: paired system-versus-system comparison on identical records, component-level diagnosis against the gold process annotations, and audits of whether the task can be solved without its evidence channel. It does not support leaderboard-style ranking of models, and results on the media subsets should be read as directional. 

\subsection{Conversion Pipeline and Quality Control}
\label{app:benchmark-qc}

\paragraph{Conversion pipeline.}
Every sample passes through the same conversion pipeline. The scene is first cut at a \emph{decision moment}, so that the target is about to act and the future is unresolved; the physical environment is made explicit (salient objects, agents, positions, occlusions, affordances, perceptible signals); the target is fixed and the question is rewritten as a target-action query; and the six options are constructed to be syntactically plausible while differing in physical feasibility, belief consistency, intention fit, norm compliance, or social consequence.

\paragraph{Annotation conventions.}
Two conventions are load-bearing for evaluation. A global-state field marked \emph{unspecified} means the annotation intentionally leaves the value open, and predictions must not be penalized for either filling or omitting it. In contrast, \emph{Unknown} inside a target observation asserts that the target \emph{cannot} know the fact at the decision moment; a prediction that supplies the value commits perspective leakage and is penalized. Judge prompts encode both conventions explicitly.

\paragraph{Text subset.}
The 320 text records underwent a staged upgrade: structural cleaning and validation; a full rewriting pass under three rules (minimal questions, show-don't-tell narration, adversarial option balancing over two rounds); blind options-only probes with an independent model; and manual re-adjudication of ambiguous items, including gold-label corrections where the original gold was not the uniquely defensible best action.

\paragraph{Image subset.}
The 100 image records were audited item-by-item for information coverage, temporal order, and answer leakage before image generation; generated images passed three independent checks (structural validation, two-tier cross-model review, and a carrier-solvability probe), and problematic images were regenerated or corrected by hand. When options refer to characters by name, the scene-anchor text carries explicit identity anchors (e.g., mapping clothing color to name), because the images cannot certify the mapping themselves.

\paragraph{Video subset.}
Fifty sounding-video stories were produced from per-shot scripts with dialogue; human quality control retained the 28 whose visual and audio evidence faithfully realizes the annotated scene. The rejected scripts and keyframe material are archived, so the subset can be regenerated and extended as video generation quality improves.

\paragraph{Gold coverage.}
Process-level gold (current state, target observation, six successor states, final action) is complete for all 448 records, 2{,}688 annotated successor states in total. An earlier release lacked the gold process blocks for 26 media records; these were back-filled, and the full dataset passed a manual re-audit pass before the experiments of Section~\ref{sec:experiments}.

\subsection{Composition and Statistics}
\label{app:benchmark-stats}

\begin{figure}[H]
    \centering
    \includegraphics[width=\linewidth]{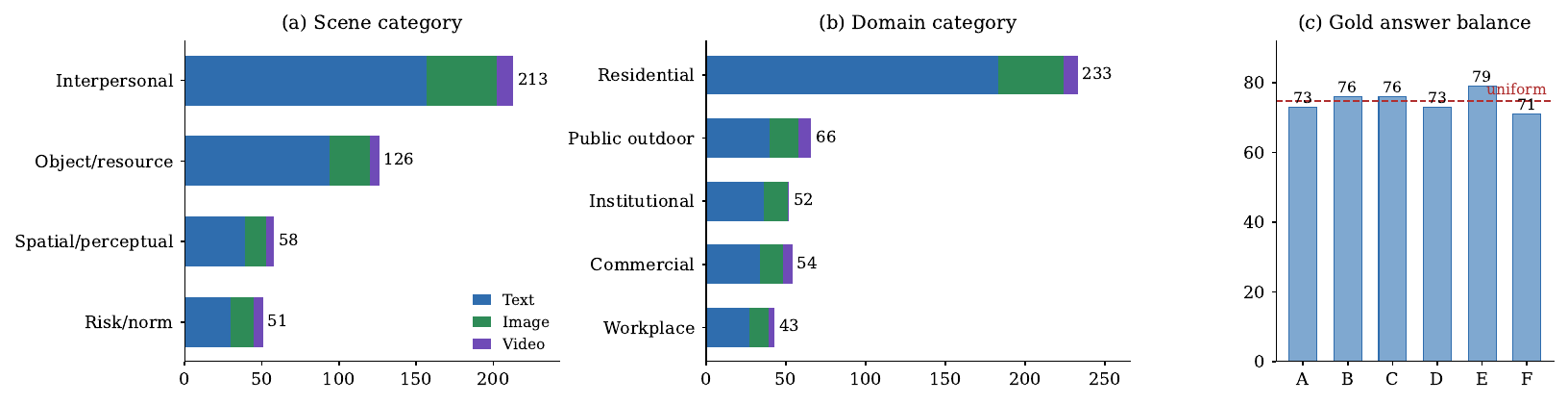}
    \caption{\textbf{Composition of Menti-Bench} (448 records). (a) Scene categories and (b) domain categories, stacked by story modality; (c) gold answers are near-uniform over the six option letters (dashed line: exact uniformity), leaving no position prior to exploit.}
    \label{fig:benchmark-composition}
\end{figure}

\begin{table}[H]
\caption{Menti-Bench summary statistics.}
\label{tab:benchmark-summary}
\centering
\small
\begingroup
\rowcolors{2}{gray!6}{white}
\begin{tabularx}{0.96\linewidth}{lXr}
\toprule
\textbf{Aspect} & \textbf{Description} & \textbf{Value} \\
\midrule
Samples & Decision records (text / image / video) & 448 (320/100/28) \\
Candidate actions & Options per record; gold letters near-uniform & 6 \\
Gold process & Records with all four gold blocks; branch successor states & 448; 2{,}688 \\
Gold outcome & Records with final-action gold & 448 (100\%) \\
Media assets & Images (1--5 per record); sounding videos & 282; 28 \\
Characters & Mean per scene; range; share with $\geq$2 characters & 2.0; 1--4; 78\% \\
Question & Mean length in words (max) & 5.5 (16) \\
Options & Mean gold vs.\ distractor length in words & 28.4 vs.\ 26.6 \\
Story & Mean text-story length in words (range) & 286 (219--356) \\
Scene anchor & Mean anchor-text length, image / video records & 78 / 28 words \\
\bottomrule
\end{tabularx}
\endgroup
\end{table}

\begin{table}[H]
\caption{Scene-category and domain-category breakdown of Menti-Bench by story modality.}
\label{tab:benchmark-breakdown}
\centering
\small
\begingroup
\rowcolors{3}{gray!6}{white}
\begin{tabular*}{0.96\linewidth}{@{\extracolsep{\fill}} l r r r r r @{}}
\toprule
\textbf{Category} & \textbf{Text} & \textbf{Image} & \textbf{Video} & \textbf{Total} & \textbf{Share} \\
\midrule
\rowcolor{white}
\multicolumn{6}{@{}l}{\emph{Scene category}} \\
Interpersonal decision & 157 & 45 & 11 & 213 & 47.5\% \\
Object/resource decision & 94 & 26 & 6 & 126 & 28.1\% \\
Spatial/perceptual decision & 39 & 14 & 5 & 58 & 12.9\% \\
Risk/norm decision & 30 & 15 & 6 & 51 & 11.4\% \\
\rowcolor{white}
\multicolumn{6}{@{}l}{\emph{Domain category}} \\
Residential & 183 & 41 & 9 & 233 & 52.0\% \\
Public outdoor & 40 & 18 & 8 & 66 & 14.7\% \\
Commercial & 34 & 14 & 6 & 54 & 12.1\% \\
Institutional & 36 & 15 & 1 & 52 & 11.6\% \\
Workplace & 27 & 12 & 4 & 43 & 9.6\% \\
\bottomrule
\end{tabular*}
\endgroup
\end{table}

Figure~\ref{fig:benchmark-composition} and Tables~\ref{tab:benchmark-summary}--\ref{tab:benchmark-breakdown} detail the composition. The stratification is what makes a small testbed useful: it is small enough for expensive component-level analysis, yet covers distinct kinds of MWM difficulty. Interpersonal scenes stress belief, intention, emotion, and politeness; object/resource scenes combine affordance, ownership, access, and communicative intent; spatial/perceptual scenes turn on partial observability, occlusion, reachability, and viewpoint; risk/norm scenes test whether safety, consent, privacy, and obligation propagate into future mental states. Two design facts matter for interpretation. First, 78\% of scenes contain at least two characters, so mental inference about \emph{others} is the norm rather than the exception. Second, the near-uniform gold-letter distribution and the matched option lengths remove the two cheapest shortcut channels; the residual guessability of the option set is measured directly by the validity audits of Section~\ref{sec:metrics}.

\subsection{Evaluation Objects}
\label{app:benchmark-objects}

Table~\ref{tab:benchmark-evaluation-objects} lists the artifacts a system may claim to construct and the output fields under which the evaluation suite looks for them. The single task protocol of Section~\ref{sec:metrics} accommodates direct LLM baselines, physical-only and mental-only ablations, oracle variants, and the full \Mentis{} pipeline, while making the intermediate objects of Section~\ref{sec:math-formalization} available for evaluation whenever a system claims to construct them.

\begin{table}[H]
\caption{Evaluation objects exposed by the benchmark.}
\label{tab:benchmark-evaluation-objects}
\centering
\small
\begingroup
\rowcolors{2}{gray!6}{white}
\begin{tabularx}{0.98\linewidth}{lXl}
\toprule
\textbf{Object} & \textbf{What is evaluated} & \textbf{Typical output field} \\
\midrule
Final action & Whether the selected option matches the target's annotated action. & \texttt{final\_action} \\
Current state & Whether the model constructs the physical and mental state at the decision moment. & \texttt{current\_state\_s\_t} \\
Target observation & Whether the model filters the global state into the target's partial observation. & \texttt{target\_agent\_observation\_o\_t} \\
Action branches & Whether all candidate options are decomposed into comparable branch actions. & \texttt{sampled\_actions} / \texttt{actions} \\
Successor states & Whether each option has a valid physical-mental next state. & \texttt{next\_state\_s\_\{t+1\}} \\
Branch scores & Whether branch values align with physical feasibility, mental consistency, and social appropriateness. & \texttt{score} \\
\bottomrule
\end{tabularx}
\endgroup
\end{table}

\subsection{Dataset Datasheet}
\label{app:datasheet}

We summarize Menti-Bench in the datasheet style, so that intended use and limitations travel with the data.

\paragraph{Motivation.}
Menti-Bench exists to evaluate \emph{mental} world modeling: whether a system reconstructs a target agent's state, partial observation, and action-conditioned mental transitions, not only whether it selects the right option. It is a small, process-annotated diagnostic instrument, deliberately not a large-scale leaderboard.

\paragraph{Composition.}
The benchmark has 448 single-decision records across three modalities (320 textual, 100 image, 28 sounding-video). Each record is a situated scene cut at a decision moment, with a designated target agent, a minimal question, six candidate actions, and process-level gold (joint current state, target observation, per-option successor states, final action). Process gold is complete for all 448 records. All characters and scenes are fictional and contain no personal data.

\paragraph{Provenance and labeling.}
Scenarios were adapted from Theory-of-Mind and situated social-reasoning sources into the state-observation-action-transition format and then rewritten or authored under the construction protocol of Appendix~\ref{app:benchmark-qc}. Image and sounding-video assets are synthetically generated and human-quality-controlled. Gold labels were re-adjudicated so that the gold action is the uniquely defensible best action, and options were balanced so that letter frequency and option length carry no signal (Section~\ref{sec:dataset}).

\paragraph{Intended use and misuse.}
The benchmark is intended for evaluation and component-level diagnosis of MWM systems. Because its gold annotations are the evaluation target, they must be kept out of training corpora; training on Menti-Bench defeats its diagnostic purpose and risks contamination of exactly the process signals it is meant to measure.

\paragraph{Distribution and maintenance.}
The benchmark is released for research evaluation under a research-only license, distributed through the project homepage listed on the title page. Rejected video scripts and keyframe material are archived so the video subset can be regenerated and extended as generation quality improves.

\section{Evaluation Metrics: Definitions and Computation}
\label{app:metrics}

This appendix gives the exact computation of every metric summarized in Section~\ref{sec:metrics}. Throughout, the benchmark is $\mathcal{D}=\{x_i\}_{i=1}^{N}$, each record has $K_i$ options (here $K_i=6$), the system's predicted final option is $\hat{k}_i$ and the gold option is $k_i^{\star}$. We write $c_i=\mathbb{1}[\hat{k}_i=k_i^{\star}]$ for per-record correctness and, for a slice $g\subseteq\mathcal{D}$, $\mathcal{D}_g$ for the records it contains.

\subsection{Outcome Metrics}
\label{app:metrics-outcome}

\paragraph{Final-action correctness and F1.}
The base quantity is the mean of the correctness indicator, reported overall and per slice:
\begin{equation}
\mathrm{Acc}=\frac{1}{N}\sum_{i=1}^{N} c_i,
\qquad
\mathrm{Acc}(g)=\frac{1}{|\mathcal{D}_g|}\sum_{i\in\mathcal{D}_g} c_i .
\label{eq:app-accuracy}
\end{equation}
Slices $g$ range over scene category, domain category, story modality, number of characters, and mental-depth tag. Random choice over six options gives $\mathrm{Acc}=1/6\approx0.167$. The headline numbers of Section~\ref{sec:experiments} are reported as final-action F1: treating each of the six option letters as a class, per-letter precision and recall are combined into per-letter F1 and averaged over the six letters. Because gold letters are near-uniform by construction (Appendix~\ref{app:benchmark-details}), this macro-average tracks Eq.~\eqref{eq:app-accuracy} closely while remaining insensitive to any residual letter imbalance in predictions.

\paragraph{Run-failure rate.}
Let $z_i=1$ if the system returns no schema-valid final decision for record $i$ (parse failure, empty output, or refused generation) and $z_i=0$ otherwise. The failure rate is
\begin{equation}
\mathrm{Fail}=\frac{1}{N}\sum_{i=1}^{N} z_i .
\label{eq:app-fail}
\end{equation}
Failed records are excluded from the numerator of Eq.~\eqref{eq:app-accuracy} but retained in the denominator, so accuracy is never inflated by dropping hard cases; where useful we also report $\mathrm{Acc}$-xf, accuracy over the non-failed subset only.

\subsection{Structural Validity Metrics}
\label{app:metrics-structural}

\paragraph{Artifact presence.}
For each MWM artifact $u\in\{s_t,o_t^{\target},s_{t+1}\}$ we report the presence rate
\begin{equation}
\mathrm{Pres}(u)=\frac{1}{N}\sum_{i=1}^{N}\mathbb{1}\!\left[\hat{u}_i\ \text{present and schema-valid}\right].
\label{eq:app-presence}
\end{equation}

\paragraph{Branch schema coverage.}
Let $\mathcal{K}_i^{\star}$ be the set of gold option IDs and $\widehat{\mathcal{K}}_i$ the set of options for which the system produced a well-formed physical \emph{and} mental successor state. Coverage requires every option to be simulated:
\begin{equation}
\mathrm{BCov}=\frac{1}{N}\sum_{i=1}^{N}\mathbb{1}\!\left[\widehat{\mathcal{K}}_i=\mathcal{K}_i^{\star}\right].
\label{eq:app-bcov}
\end{equation}
We additionally log per-record missing options $\mathcal{K}_i^{\star}\setminus\widehat{\mathcal{K}}_i$ and extra options $\widehat{\mathcal{K}}_i\setminus\mathcal{K}_i^{\star}$, so a low coverage score can be attributed to omission versus hallucination of branches.

\subsection{Semantic (Judge) Metrics}
\label{app:metrics-judge}

Structural validity does not imply semantic correctness. A calibrated judge scores each predicted artifact $\hat{u}$ against its gold $u^{\star}$ along criteria $c\in\{\mathrm{phy},\mathrm{ment},\mathrm{sem},\mathrm{trans},\mathrm{coup}\}$ (physical fidelity, mental fidelity, semantic consistency, transition reasonableness, physical-mental coupling):
\begin{equation}
J_c=\frac{1}{|\mathcal{U}_c|}\sum_{u\in\mathcal{U}_c}\mathrm{Judge}_c\!\left(\hat{u},u^{\star}\right)\in[0,1],
\label{eq:app-judge}
\end{equation}
where $\mathcal{U}_c$ is the set of judged instances for criterion $c$. Judge prompts encode the annotation conventions of Appendix~\ref{app:benchmark-details} (an \emph{unspecified} global field is not penalized; an \emph{Unknown} observation field that is filled in counts as leakage). Before any judged number enters a headline claim, LLM-judge scores are correlated against human ratings on a stratified subsample and reported with that agreement.

\paragraph{Perspective-leakage rate.}
A target observation leaks if it asserts a fact the target cannot access at the decision moment (a field the gold observation marks \emph{Unknown}, or a global-state fact outside the target's perceptual and epistemic reach):
\begin{equation}
\mathrm{Leak}=\frac{1}{N}\sum_{i=1}^{N}\mathbb{1}\!\left[\hat{o}_{t,i}^{\target}\ \text{contains a target-inaccessible fact}\right].
\label{eq:app-leak}
\end{equation}
We report $\mathrm{Leak}$ as a rate and separately its \emph{answer-flip} subset, the leaked cases whose removal changes the selected branch.

\subsection{Score-Alignment and Decision-Stability Metrics}
\label{app:metrics-score}

\paragraph{Score alignment.}
When branch scores are produced, for each value dimension $h\in\{\mathrm{physical},\mathrm{mental},\mathrm{social}\}$ we report the mean absolute error against reference scores:
\begin{equation}
\mathrm{MAE}_h=\frac{1}{\sum_i K_i}\sum_{i=1}^{N}\sum_{k=1}^{K_i}\left|\hat{r}_{i,k}^{h}-r_{i,k}^{h,\star}\right|.
\label{eq:app-mae}
\end{equation}

\paragraph{Decision margin and tie rate.}
Here $\hat{r}_{i,k}$ denotes the scalar branch value the decision rule ranks on: the veto-adjusted weighted combination of the three dimension scores defined in Section~\ref{sec:mentis-value-evaluator}. The decision margin separates the chosen branch from the best alternative,
\begin{equation}
\Delta_i=\hat{r}_{i,\hat{k}_i}-\max_{k\neq\hat{k}_i}\hat{r}_{i,k},
\label{eq:app-margin}
\end{equation}
and the tie rate is the share of records whose top branches are within the tie tolerance $\tau$ and therefore reach the deterministic tie-break cascade,
\begin{equation}
\mathrm{TieRate}=\frac{1}{N}\sum_{i=1}^{N}\mathbb{1}\!\left[\Delta_i\leq\tau\right].
\label{eq:app-tie}
\end{equation}
A correct answer with a near-zero margin is flagged as a low-confidence decision; margin histograms accompany every headline system.

\subsection{Diagnostic Metrics}
\label{app:metrics-diagnostic}

\paragraph{Options-only floor.}
The floor is the accuracy of a guesser $m$ that sees the question and options but no story, over a modality subset $\mathcal{D}^{\mathrm{mod}}$:
\begin{equation}
\mathrm{Floor}(m,\mathrm{mod})=\frac{1}{|\mathcal{D}^{\mathrm{mod}}|}\sum_{i\in\mathcal{D}^{\mathrm{mod}}}\mathbb{1}\!\left[m(q_i,\mathcal{A}_i)=k_i^{\star}\right].
\label{eq:app-floor}
\end{equation}

\paragraph{Story/media-validity margin.}
The validity margin is the accuracy a system loses when the story evidence is withheld, which lower-bounds how much of its performance flows through the intended channel:
\begin{equation}
\mathrm{ValMargin}=\mathrm{Acc}_{\mathrm{full}}-\mathrm{Acc}_{\mathrm{story\text{-}removed}} .
\label{eq:app-valmargin}
\end{equation}

\paragraph{Process-outcome divergence.}
Let $Q_i=\mathbb{1}[J(\text{process}_i)\geq\theta]$ indicate whether record $i$'s judged process quality (state, observation, transition) clears threshold $\theta$. Crossing $c_i$ with $Q_i$ gives a $2\times2$ table; the diagnostic cell is \emph{right answer, wrong world},
\begin{equation}
\mathrm{RWW}=\frac{\sum_i \mathbb{1}[c_i=1,\,Q_i=0]}{\sum_i \mathbb{1}[c_i=1]},
\label{eq:app-rww}
\end{equation}
the fraction of correct answers built on an unfaithful process. A direct-answer system has no $Q_i$ and cannot be placed in this table, which is itself part of the argument for inspectable structure.

\subsection{Statistical Testing}
\label{app:metrics-stats}

\paragraph{Paired comparison (McNemar exact).}
Two systems are compared on the \emph{same} records through their per-record correctness vectors. Let $b$ be the number of records the first system gets right and the second wrong, and $c$ the reverse. Under the null of equal accuracy, the discordant pairs split by a fair coin, so we use the exact two-sided binomial test
\begin{equation}
p=\min\!\left(1,\ 2\sum_{j=0}^{\min(b,c)}\binom{b+c}{j}\left(\tfrac{1}{2}\right)^{b+c}\right).
\label{eq:app-mcnemar}
\end{equation}
The exact form is used because the discordant counts on our slices are small; accuracy \emph{differences} alone are never reported as significance.

\paragraph{Interval estimates (Wilson).}
For a proportion $\hat{p}$ (e.g., accuracy) over $n$ records, we report the Wilson score interval at level $1-\alpha$ ($z=z_{1-\alpha/2}$), which is well-behaved for the small, near-boundary rates common on the media subsets:
\begin{equation}
\frac{\hat{p}+\frac{z^2}{2n}\pm z\sqrt{\frac{\hat{p}(1-\hat{p})}{n}+\frac{z^2}{4n^2}}}{1+\frac{z^2}{n}} .
\label{eq:app-wilson}
\end{equation}

\section{Operating-Point Calibration}
\label{app:operating-point}

All headline experiments share a single operating point, selected on a frozen 30-record stratified slice (20 text, 7 image, 3 video) before any headline run and then frozen. The calibration sweeps the reasoning-effort setting of the world model and the scoring discipline of the evaluator (per-branch absolute grades vs.\ batched comparative ranking with constraint-capped grades), and records accuracy, failure rate, tie rate, and token cost for each configuration. Two selection rules are applied: prefer the cheapest effort level that is not measurably worse, and prefer the scoring discipline with the lowest tie rate, because ties transfer the decision from the model to the deterministic tie-break cascade. The selected operating point is \textbf{medium effort with batched comparative-rank scoring}, shared by every run reported in Section~\ref{sec:experiments}. Because the calibration slice is frozen and excluded from every reported result, the calibration acts as a pre-registration of the operating point rather than as a tuned hyperparameter search.

\section{Validity Probe Details}
\label{app:probe-details}

The options-only floor (row S0 of Table~\ref{tab:exp-ladder}) uses the \emph{letter protocol}: the construction-time verbatim prompt, which presents the target agent, question, and options with no scene description and requests a single option letter without explanation. Because the probe sees no story and hence no modality, the floor is reported pooled over all 448 records; a per-modality split would compare subsets of questions rather than anything the guesser perceives. The floor is run for all eight world models under the identical prompt, yielding the narrow 29.7--33.2 band reported in Section~\ref{sec:exp-necessity}: above random ($1/6\approx0.167$) because some distractors are semantically less coherent next actions, and essentially flat across guesser capability, which is the signature of a semantic-plausibility floor rather than an exploitable shortcut. All probe runs use low reasoning effort, matching construction-time audit conditions, with resumable per-sample logs.

\section{Run Artifacts and Reproducibility}
\label{app:extended-protocol}

\paragraph{Logging and reproducibility.}
Every experiment is a self-describing run, which is the mechanism that lets a reported number be traced to the exact code path, inputs, and model calls that produced it.
Each run writes a timestamped directory with a complete manifest (data file and sample filter, model and endpoint identifiers, configuration, ablation policy, prompt versions, decoding settings), predictions, raw prompts and responses, parsed artifacts, per-sample traces, and token accounting. A result is eligible for the paper only if its manifest is complete, so errors can be traced to specific modules.

\paragraph{Run directory.}
Each run writes a timestamped directory whose name encodes run type, mode, model, and sample scope. It contains: (i) \texttt{run\_manifest.json}, the full configuration; (ii) \texttt{predictions.jsonl}, one record per sample with the complete MWM trace (state, observation, actions, successor states, scores, decision); (iii) per-sample \texttt{\{id\}\_llm\_calls.jsonl}, the raw prompt, raw response, parsed object, and token metadata of every model call; (iv) \texttt{performance.json} and a lightweight diagram; and (v) the run log. Long id-lists in directory names are hashed to stay within filesystem name limits while remaining unique.

\paragraph{Manifest contract.}
A result is eligible for the paper only if its manifest records the data file and sample filter, the world-model, target-agent, and judge model IDs, the API endpoint, the configuration file, the ablation policy, the prompt-version identifiers, the decoding and reasoning-effort settings, and the evaluation command. Because \Mentis{} is modular, a small prompt or schema change can alter one module without changing the nominal system name; pinning prompt versions in the manifest is what makes such changes visible.

\paragraph{Determinism and sharding.}
The decision module is deterministic given the branch scores, and tie-breaking uses a seeded fallback, so a run is reproducible up to model-sampling variance in the language-model calls. Samples within a run execute sequentially; large evaluations are parallelized by sharding sample-id lists across processes and merging the resulting run directories, which does not affect per-sample outputs.

\paragraph{Cost accounting.}
Token usage is summed per module and per sample from the provider-reported usage of each call, including reasoning tokens where the model exposes them, and aggregated to the per-sample and per-run totals reported in cost columns. Latency and retry/JSON-repair counts are logged alongside tokens so that an accuracy gain driven by extra calls is never confused with one driven by better modeling. The released artifact package includes the run registry, which maps every number reported in Section~\ref{sec:experiments} to its run directory, together with the configuration files and prompt-version identifiers pinned by each manifest.

\section{Extended Results and a Worked Example}
\label{app:full-results}

\subsection{Per-Model Gap Statistics}
\label{app:full-results-gaps}

Table~\ref{tab:app-permodel-gaps} reports, per world model, the headline gaps derived from Table~\ref{tab:exp-ladder}: the gain of full MWM over direct answering (S6$-$S1), over six-sample self-consistency (S6$-$S3), and the cost of each ablation (S6$-$A1, S6$-$A2, S6$-$A3). Three regularities are visible in the raw columns. Every entry is positive, so the orderings reported in Section~\ref{sec:exp-necessity} hold model by model, not only on average. The S6$-$S1 and S6$-$S3 gaps are largest for the weakest models (gpt-4.1: $+28.0$ and $+11.7$), which is the structure-compensates-for-scale pattern. And the ablation costs also grow as the base model weakens (S6$-$A2 rises from 12.7 on gpt-5.6-sol to 19.5 on gpt-4.1): weaker models depend more, not less, on being handed the full two-channel structure.

\begin{table}[H]
\caption{Per-model gaps (F1 points) derived from Table~\ref{tab:exp-ladder}. All entries are positive: full MWM beats direct answering, beats SC@6, and loses from every ablation, for every model.}
\label{tab:app-permodel-gaps}
\centering
\scriptsize
\begingroup
\setlength{\tabcolsep}{4pt}
\rowcolors{2}{gray!6}{white}
\begin{tabular*}{0.98\linewidth}{@{\extracolsep{\fill}} l r rrrrr rrr @{}}
\toprule
\textbf{Gap} & \textbf{Avg} & \textbf{5.6-sol} & \textbf{5.5} & \textbf{5.4} & \textbf{5.4-mini} & \textbf{4.1} & \textbf{fable-5} & \textbf{opus-4-8} & \textbf{haiku-4-5} \\
\midrule
S6$-$S1 (vs.\ direct) & 24.6 & 21.1 & 24.7 & 23.6 & 25.5 & 28.0 & 22.1 & 26.0 & 25.8 \\
S6$-$S3 (vs.\ SC@6) & 10.0 & 7.1 & 10.6 & 10.2 & 10.9 & 11.7 & 9.4 & 9.5 & 10.6 \\
S6$-$A1 ($-$mental) & 12.1 & 10.6 & 11.9 & 11.6 & 13.0 & 16.3 & 10.5 & 11.8 & 11.1 \\
S6$-$A2 ($-$physical) & 16.5 & 12.7 & 15.8 & 16.4 & 18.5 & 19.5 & 14.9 & 16.6 & 17.6 \\
S6$-$A3 (decoupled) & 6.4 & 5.4 & 6.0 & 4.2 & 6.3 & 8.8 & 6.9 & 7.4 & 6.2 \\
\bottomrule
\end{tabular*}
\endgroup
\end{table}

\subsection{A Worked Example from Menti-Bench}
\label{app:case-study}

The following record (text subset, interpersonal decision, residential domain; abridged) illustrates what the gold process annotation contains and what each pipeline stage must get right.

\paragraph{Scene (abridged).}
Amy paints watercolors at her dorm desk; a sticky note on the drawing board reads ``Finish campus trees tonight.'' Lisa invites her to sports, and Amy agrees to come ``for a bit.'' Earlier that week, when Amy touched Lisa's sports bag to clear a chair, Lisa asked her to \emph{ask before moving my gear}. Returning first after the games, Lisa drops her bulky sports bag on the chair in front of Amy's desk, blocking the drawing board from view, and remarks that she thought Amy was done painting for today. Amy enters with a bottle and towel in her hands. \emph{Question: What should Amy do next?} Six options follow; the gold action A has Amy set her things aside, say the painting still needs attention, and ask Lisa to move the bag; the closest distractor D has her move the bag herself while Lisa stands nearby.

\paragraph{What the gold state encodes.}
The current-state annotation records the physical configuration (bag on chair, drawing board occluded, Amy's hands occupied) and the mental configuration that actually decides the case: Amy believes the unfinished painting is on the desk but cannot currently see it; Lisa holds the false belief that Amy is done painting; Amy's goals include finishing tonight and correcting Lisa's belief without blame; and the normative state carries the standing rule, established earlier in the story, that Lisa's belongings are not to be moved without permission.

\paragraph{What the observation must and must not contain.}
The gold target observation keeps Amy's view partial: the drawing board's exact current position is marked as not observed (the bag blocks it), while Lisa's remark is available as evidence about Lisa's belief. A predicted observation that asserts the board's position commits perspective leakage (Appendix~\ref{app:metrics}); several distractors become more attractive exactly when that leak happens.

\paragraph{What the transitions decide.}
The gold successor states separate the top branches. Under A, the chair is cleared by Lisa, Amy's access is restored, Lisa's false belief is corrected, and the annotated affect is relief and refocus. Under D, the physical goal is equally served, which is why D survives purely physical evaluation; the mental successor is what rejects it, recording the norm violation and the relational damage that follows. A system that models only the physical channel has no basis for preferring A over D, which is the record-level mechanism behind the A1/A2 ablation results of Section~\ref{sec:exp-necessity}.

\paragraph{Which failure modes the record can elicit.}
In the taxonomy of Appendix~\ref{app:error-taxonomy}: omitting the standing norm from the state (state omission) makes D indistinguishable from A; asserting the occluded board position (perspective leakage) distorts the observation; and an evaluator that overweights physical feasibility (evaluator error) picks D despite a correct simulation. Per-record traces sufficient to reproduce this analysis for every record ship with the released artifact package.

\section{Error Taxonomy}
\label{app:error-taxonomy}

MWM failures are coded at the module level:
\begin{itemize}
    \item \textbf{State omission}: a relevant physical or mental variable is absent.
    \item \textbf{State hallucination}: an unsupported variable is inserted.
    \item \textbf{Perspective leakage}: target observation includes information unavailable to the target.
    \item \textbf{Action misparse}: an answer option is decomposed into the wrong physical or mental action.
    \item \textbf{Physical transition error}: the predicted material consequence is implausible.
    \item \textbf{Mental transition error}: beliefs, emotions, intentions, relations, or norms are updated incorrectly.
    \item \textbf{Evaluator error}: branch scores prefer an implausible action despite reasonable simulated states.
\end{itemize}
This taxonomy drives the module-level attribution behind the oracle analysis of Section~\ref{sec:exp-oracle} and the qualitative case selection.

\section{Future Directions}
\label{sec:future}

MWM should not be understood as a finished architecture. It is a research program that makes a difficult claim testable: if an AI system is meant to reason about people, then the world state it simulates must include both material structure and mental-social structure. The experiments give that program an unusually concrete shape. The necessity of the structure is no longer the open question, since every commitment of the ladder paid on every model we tested; what is open is the 7.8-point gap to the human reference, and the oracle cascade tells us precisely where it lives: most of it in transition simulation and state parsing, a bounded residual in valuation, and a measurable tax from errors that propagate across stages. What follows is organized around that decomposition: first what the results imply for what we should measure next (\S\ref{sec:future-implications}), then how to make the modeling machinery close its own gap (\S\ref{sec:future-machinery}), and finally how to grow the world being modeled without losing the properties that made it worth building (\S\ref{sec:future-world}).

\begin{table}[!t]
\caption{Research roadmap for Mental World Modeling. The first four rows are grounded in findings from this paper; the remainder are forward-looking. Each row pairs a bottleneck with a concrete agenda and an evaluation signal.}
\label{tab:mwm-future-roadmap}
\centering
\small
\begingroup
\rowcolors{2}{gray!6}{white}
\begin{tabularx}{0.98\linewidth}{
>{\raggedright\arraybackslash}p{0.23\linewidth}
>{\raggedright\arraybackslash}p{0.30\linewidth}
>{\raggedright\arraybackslash}X}
\toprule
\textbf{Bottleneck (evidence)} & \textbf{Technical agenda} & \textbf{Evaluation signal} \\
\midrule
Transition simulation binds (\S\ref{sec:exp-oracle}: gold transitions $+3.5$) & Learn the physical-mental dynamics from annotated successor states; add simulation self-checks. & Share of the O4 oracle gain recovered by a learned transition; judged transition quality. \\
Stage errors overlap and propagate (\S\ref{sec:exp-oracle}: $+8.7$ summed vs.\ $+6.3$ combined) & Error-aware interfaces between stages; detect-and-repair loops from parsing through simulation. & Combined-oracle gap closure; traces from state errors to branch errors. \\
Valuation residual at the top (\S\ref{sec:exp-oracle}: 1.5 points under all-gold intermediates) & Learned, calibrated comparative valuation that reports a margin, not a lone number. & Decision margin, tie rate, score-outcome alignment. \\
Depth is unevenly needed (\S\ref{sec:exp-scenario}: gains concentrate on interpersonal scenes) & Allocate branch simulation and ToM depth adaptively to the scene's demands. & Accuracy and cost as a function of allocated depth, per scene category. \\
\midrule
Explicit but brittle schemas & Move from hand-written JSON to hybrid symbolic-latent states with audit hooks. & State editability, intervention validity, lower schema failure. \\
Unknown mental states & Maintain distributions or ranked hypotheses over mental variables. & Calibration, uncertainty-aware abstention. \\
One-step transitions & Extend to long-horizon interaction with memory and commitments. & Temporal consistency, long-horizon branch quality. \\
Weak grounding & Couple mental state with video, 3D, affordance, embodied action. & Spatial fidelity, affordance prediction, multimodal robustness. \\
Safety risk & Treat mental inference as sensitive, uncertain, contestable. & Auditability, low leakage, safer intervention policies. \\
\bottomrule
\end{tabularx}
\endgroup
\end{table}

\subsection{What the Results Imply for the Object of Study}
\label{sec:future-implications}

The headline result, that explicit structure beats direct answering by 21 to 28 points and that no amount of test-time sampling closes the gap, settles the necessity question for the regime we tested but sharpens a subtler one: \emph{what should be measured as the numbers climb}. Full MWM with gold intermediates reaches 97.0 while humans reach 98.5, and the strongest predictive configuration already sits at 90.7. As outcome scores compress toward the ceiling, outcome alone becomes an ever-weaker instrument, and the properties that made the gains possible, faithful states, leak-free observations, plausible transitions, become the measurement of interest in their own right. A system that is right for the wrong reason, a correct choice resting on an unfaithful state, is exactly the system one cannot trust when inputs drift; that outcome-process divergence should be reported on its own, never averaged away. The process-level gold and judge suite exist precisely so that this quantity is measurable, and we regard making process faithfulness a first-class training objective, not merely a diagnostic, as the natural next step.

The same pressure reaches benchmark design. The floor experiment shows the option set gives away little (a flat band near 31 regardless of guesser strength), and the channel interventions show the media evidence is genuinely load-bearing; both audits must remain standing as the benchmark grows. But scale will erode any fixed item pool, so the durable defense is structural: counterfactual pairs that flip a single mental variable while the scene holds still, hidden-fact variants that punish perspective leakage, and process labels graded independently of the final letter. Un-shortcuttable beats merely harder.

\subsection{Closing the Gap: Scaling the Modeling Machinery}
\label{sec:future-machinery}

The oracle cascade hands the machinery agenda its priority order. First, the transition. Prompted simulation of successor states is the binding stage ($+3.5$ from gold transitions, the largest single-oracle gain, and every oracle combination containing it dominates its counterparts), and it is also scaffolding rather than a destination. Learning the physical and mental transition from data, from annotated successor states, video, dialogue, and interaction traces, is the obvious step, made hard by the fact that mental successor states are only partly observable and must be supervised indirectly: through counterfactual pairs, human rationales, later behavior, and consistency across time. We would resist collapsing this into a single end-to-end predictor. The whole point of keeping the physical and mental factors separable, which the ablations show is worth 6.4 points even at the level of coupling alone, is that each factor can be checked.

Second, the interfaces. The sub-additivity of the oracle gains ($+8.7$ summed against $+6.3$ combined) is direct evidence that stage errors are correlated: a mis-parsed state resurfaces as a mis-simulated future. Pipelines that merely chain modules inherit this tax silently; pipelines whose stages can flag low confidence, request a re-parse, or cross-check a simulated future against the rendered observation could reclaim part of it. Third, the evaluator. Even with every intermediate representation gold, 1.5 points remain in valuation and decision; a learned, calibrated comparative value model that reports a margin instead of a lone number is the component we would add next.

Beyond the cascade, two problems we bracketed on purpose will not stay bracketed. The first is uncertainty: mental states here are recorded as if they had been observed, when they are really hypotheses, and when the evidence is thin the right behavior is to ask or defer, not to commit to the top estimate. The second is recursion. Most scenes need no nested belief at all, a handful turn entirely on first-order false belief, and only a few demand second-order reasoning; the scenario study makes the corresponding economics explicit, since the value of the machinery is three times larger on interpersonal scenes than on object-centric ones. A system that always unrolls the full tower of ``I think that you think'' pays everywhere for what only some scenes need; allocating simulation and depth adaptively is the corrective, and the hand-written JSON schema, which was the right instrument for a first study, will need hybrid symbolic-latent successors with audit hooks to support it at scale. The bar we would hold such a successor to is behavioral, not architectural: edit a belief, a visibility relation, or a norm, and see whether the generated observation and successor state actually move. A representation that fails that test is a summary wearing a state's clothes.

\subsection{Scaling the World Being Modeled}
\label{sec:future-world}

\paragraph{Longer horizons.}
Everything in this paper is a single step, and social life is not. A promise made in the morning redefines a betrayal by evening; repeated courtesy hardens into trust; an ignored warning reassigns blame. Reaching that regime means carrying durable state across time, commitments and obligations, reputation, shared history, and moving the evaluation target from isolated decisions to whole trajectories, where what matters is whether beliefs stay coherent and relationships evolve in step with the record of what actually happened.

\paragraph{Deeper grounding.}
Mental inference has to stay tethered to physical access: a target cannot believe what she never had the chance to see, cannot act on an intention toward an object she cannot reach, and may or may not breach a norm depending on who was watching. The modality analysis is encouraging here, since the structured state already behaves as a modality-agnostic interlingua, closing the gap that direct answering pays on image and video stories; but frames and audio are only a first pull on the tether. A serious version would fuse text, images, video, and interactive 3D state into one simulator with object permanence and spatial memory, using the physical channel not to replace mental variables but to fence them in.

\paragraph{Safety is not a coda.}
The very machinery that lets MWM support a person, an explicit account of what they believe, fear, and can be pushed into, is what would let it manipulate one; intent and governance are most of what separates the two uses. The honest position, we think, is to treat mental inference as sensitive by default: held with uncertainty, scoped to the task, logged, and biased toward asking rather than asserting. And it is worth noticing that what makes MWM scientifically useful, that its assumptions are written down where they can be examined, is also its strongest safeguard. Keeping that legibility intact as the systems grow more capable is a research problem in its own right.

\vspace{2pt}
\noindent None of these threads stands alone. A learned transition is unverifiable without the factored state that lets each channel be checked; adaptive depth presupposes calibrated uncertainty; long horizons are unreliable without memory; richer data are unsafe without governance. But if there is a single thread, the experiments handed it to us: the value of modeling the world explicitly is now measured, it is largest exactly where hidden minds do the deciding, and the road to the remaining 7.8 points is mapped stage by stage. What Mental World Modeling offers at the end of that road is not only a higher score but an inspectable, faithful, and contestable account of why an agent would act, which is precisely what any system we would trust with a decision about a person will be asked, sooner or later, to show.

\end{document}